\documentclass[11pt]{article}
\usepackage[margin=1in]{geometry}
\usepackage{times}
\usepackage[round]{natbib}

\usepackage{graphicx}
\usepackage{hyperref}
\usepackage{url}
\usepackage{caption}
\usepackage{amsmath}
\usepackage{amsfonts}
\usepackage{booktabs}
\usepackage{multirow}
\usepackage{tikz}
\usepackage{xcolor}

\definecolor{motionBlue}{RGB}{84, 145, 210}
\definecolor{contextGreen}{RGB}{91, 177, 145}
\definecolor{qualityOrange}{RGB}{224, 151, 78}

\newcommand{\ind}[1]{\mathbf{1}\!\left[#1\right]}

\title{Preserve, Reveal, Expand: Towards Faithful 4D Video Editing with Region-Aware Conditioning and Benchmarking}

\author{%
Zhangchi Hu\textsuperscript{1,2},
Wenzhang Sun\textsuperscript{2,\dag},
Xiangchen Yin\textsuperscript{1},
Jiahui Yuan\textsuperscript{1},\\
Chunfeng Wang\textsuperscript{2},
Hao Li\textsuperscript{2},
Kun Zhan\textsuperscript{2},
Xiaoyan Sun\textsuperscript{1,*}\\[0.5em]
\small \textsuperscript{1}University of Science and Technology of China \quad
\textsuperscript{2}Li Auto Inc.\\
\small \textsuperscript{\dag}Project leader \quad
\textsuperscript{*}Corresponding author\\
\small \texttt{huzhangchi@mail.ustc.edu.cn}
}
\date{}

\begin{document}

\maketitle

\begin{figure}[h]
    \centering

    \includegraphics[width=0.99\textwidth]{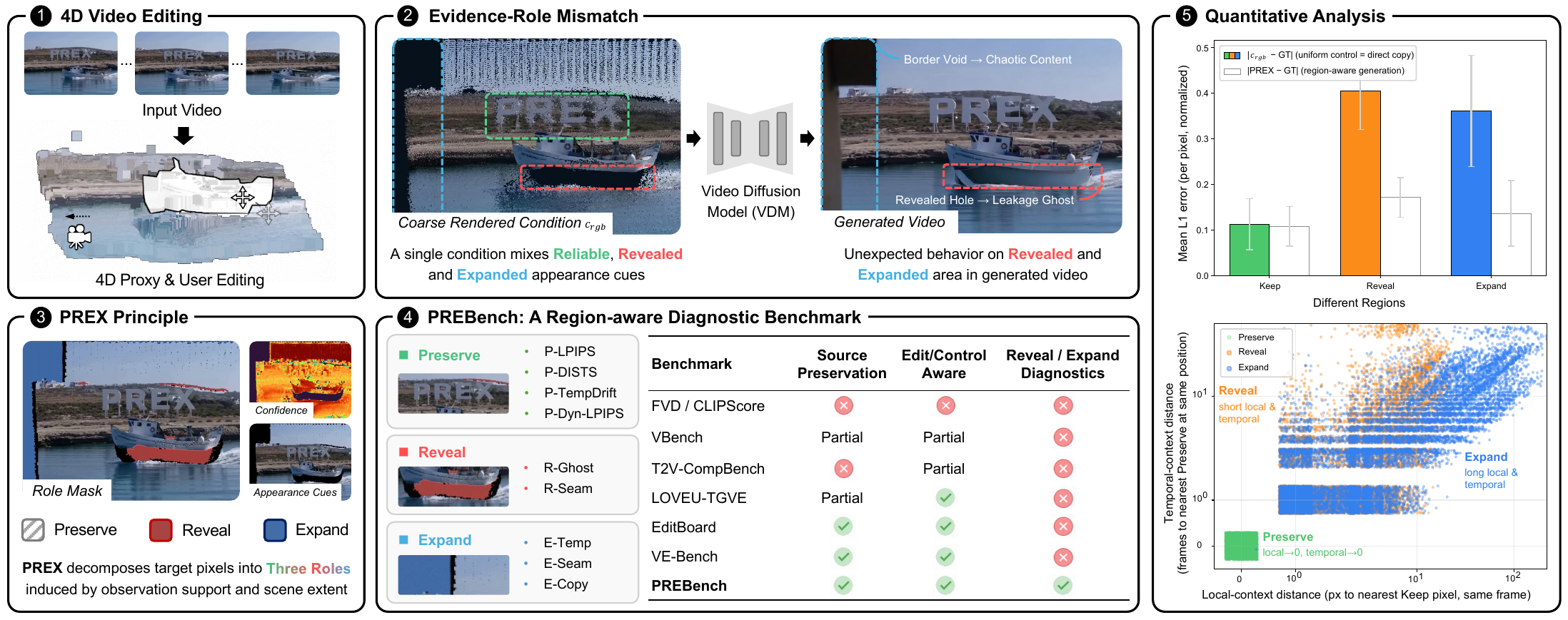}

    \caption{In 4D-driven video editing, coarse conditioning can cause the \emph{Evidence-Role Mismatch} problem. 
    PREX separates Preserve, Reveal, and Expand regions, building observation-backed cues and confidence maps to mitigate this issue. 
    Our proposed PREBench evaluates the resulting edits with region-aware metrics to diagnose such mismatches.}
    \label{fig:teaser}
\end{figure}

\begin{abstract}
While 4D-driven video diffusion models can generate visually plausible videos, faithful 4D editing poses a stricter requirement: source-observed content should be preserved, while disoccluded and out-of-view regions must be synthesized naturally. We identify a key failure mode in existing methods, which we term \emph{Evidence-Role Mismatch}: source-backed observations, uncertain rendered cues, and unsupported regions are entangled within a single conditioning signal, which can lead to preservation drift, ghosting, and unstable extrapolation. To mitigate this issue, we propose \textbf{PREX} (\textbf{Pr}eserve, \textbf{Re}veal, \textbf{Ex}pand), a region-aware framework that decomposes the target spatiotemporal volume into three distinct roles based on observation support and scene extent. PREX traces edited 4D points back to source frames and retrieves observed colors and pairs these appearance cues with geometry-aware confidence and region roles, and injects them into a frozen video diffusion backbone through an adapter trained with self-supervised proxy editing tasks. We further introduce \textbf{PREBench}, a diagnostic benchmark specifically designed for 4D video editing, to expose and quantify such region-specific failures. PREBench separately evaluates Preserve fidelity, Reveal quality, and Expand plausibility of videos generated by 4D driven video diffusion models. Experiments show that PREX substantially reduces region-specific failure modes while maintaining strong visual quality and 4D edit control compared with existing 4D-based video diffusion methods. Code will be publicly released.

\noindent\textbf{Project Page:} \url{https://ricepastem.github.io/PREX-Open}
\end{abstract}

\addtocontents{toc}{\protect\setcounter{tocdepth}{-1}}

\section{Introduction}
Recent advances in dynamic 3D reconstruction~\citep{wu2024recent, kerbl20233d, karaev2024cotracker, xiao2024spatialtracker}, 4D scene representations~\citep{wu20244d, yao2025uni4d, lu2025trackingworld, liu2025trace}, and video diffusion models~\citep{wan2025wan, yang2024cogvideox} have enabled 4D scenes to serve as structured controls for video synthesis. By lifting a monocular video into a spatiotemporal representation, 4D-driven video diffusion models can generate videos that follow prescribed camera motion, object trajectories, or scene geometry. These methods demonstrate strong controllability and visual plausibility, suggesting that 4D representations provide an effective interface between dynamic scene understanding and video generation.

\begin{figure}[t]
  \centering
  \includegraphics[width=0.95\textwidth]{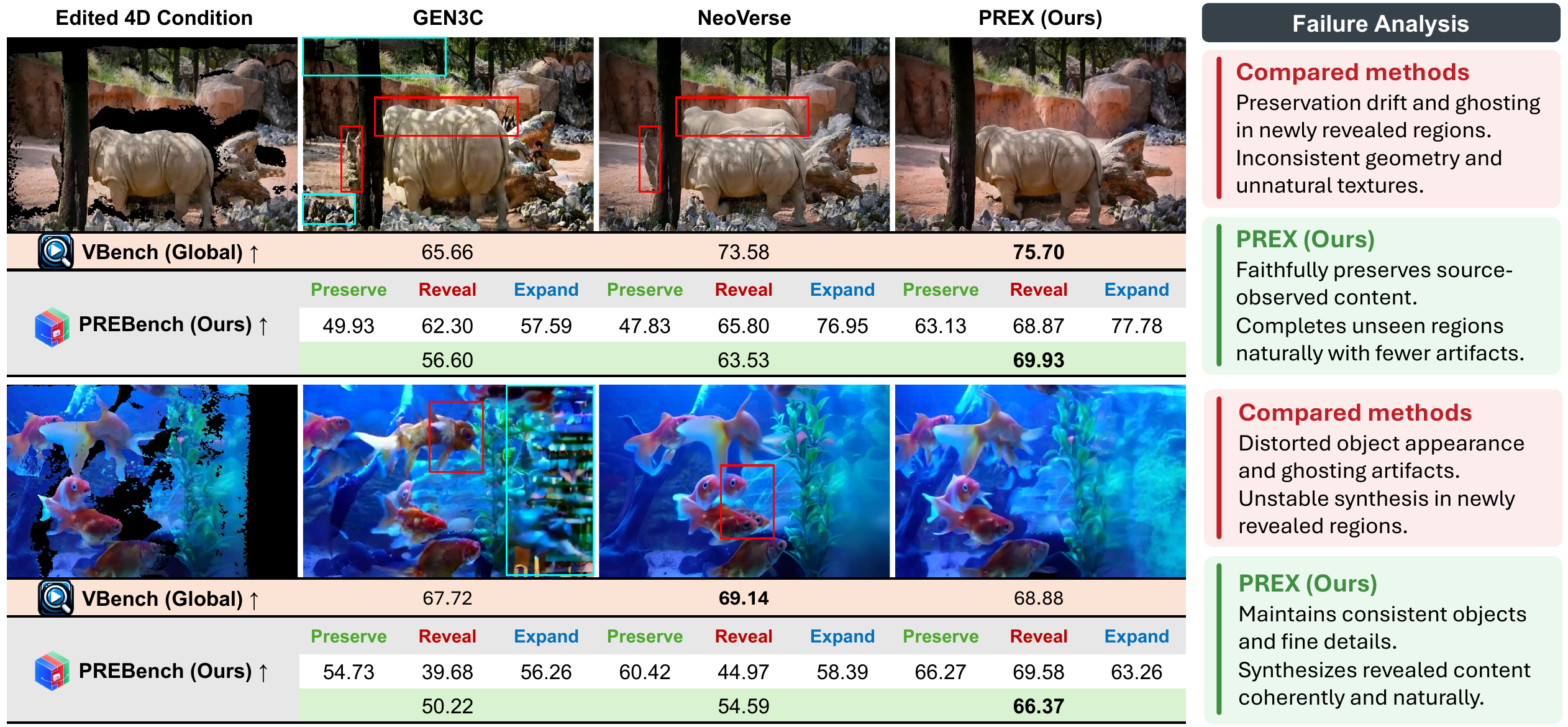}
  \caption{
    Global metrics can obscure localized failures in 4D editing.
    \textbf{PREBench} exposes these failures through region-aware evaluation of Preserve, Reveal, and Expand regions.
  }
  \vspace{-1em}
  \label{fig:new_teaser}
\end{figure}


However, faithful 4D-driven video editing imposes requirements that go beyond traditional 4D-conditioned video generation~\citep{gu2025diffusion, zheng2026versecrafter, zhang2025flextraj}. In generation, a reconstructed 4D scene can serve as geometric or motion guidance for producing a plausible target video. In editing, the input video provides appearance evidence that should be preserved whenever it remains valid under the edited camera and object motion. Thus, target regions backed by source observations should retain the identity, texture, and temporal details of the video, while disoccluded in-scene regions and newly exposed out-of-view areas must be synthesized naturally.

A key challenge is deciding when 4D evidence should be trusted or ignored. As shown in Fig.~\ref{fig:teaser}(2), pixels may correspond to reliable observations, disoccluded in-scene regions, or out-of-view areas after a 4D edit. Uniform treatment conflates distinct conditioning roles. We term this issue \textbf{Evidence-Role Mismatch}: a single conditioning signal entangles reliable source-backed evidence, uncertain rendered cues, and unsupported regions requiring generation. Thus, diffusion models must infer evidence reliability and editing role, often causing preservation drift, disocclusion ghosting, and unstable extrapolation in expanded views. Fig.~\ref{fig:teaser}(5) confirms this mismatch’s measurability. On 20 held-out cases, the coarse rendered condition has substantially higher reconstruction error in Reveal and Expand regions than Preserve regions. They exhibit distinct local-temporal context regimes: \textit{Preserve} pixels have direct observation support, \textit{Reveal} often use nearby spatiotemporal context, and \textit{Expand} typically require extrapolation over larger spatiotemporal context gaps. This motivates region-dependent conditioning for faithful editing. In particular, geometric coverage alone does not imply valid source-observation support, and in-scene unsupported pixels differ in completion role from those outside it. Existing rendered controls rarely expose both to the generator. PREX makes them explicit in conditioning representation, proxy-task curriculum, and evaluation protocol.

Motivated by this, we propose \textbf{PREX} (\textbf{Pr}eserve, \textbf{Re}veal, \textbf{Ex}pand), a framework for faithful 4D video editing. PREX decomposes the target spatiotemporal volume into \textit{Preserve} regions backed by valid source observations, \textit{Reveal} regions that are unsupported but within the original scene extent, and \textit{Expand} regions that correspond to visible out-of-view content. For each target pixel, PREX traces the edited 4D point to candidate source frames and retrieves color from a source observation that passes multiple checks and assigns confidence according to geometric support and view agreement. A region-aware adapter injects these cues and the Preserve, Reveal, and Expand roles into a frozen video diffusion backbone. To evaluate these role-specific behaviors, we introduce \textbf{PREBench}, a diagnostic benchmark with editing cases, region-role masks, region-specific metrics supported by a human preference study. Experiments show that PREX mitigates the Evidence-Role Mismatch problem while maintaining quality and 4D edit control compared to existing methods.

\vspace{-0.5em}

\section{Related Work}

\noindent\textbf{Geometry-aware Video Generation.} Recent video diffusion models incorporate camera poses, depth, point clouds, trajectories, or Gaussian-based scene representations for controllable generation~\citep{he2024cameractrl,wang2024motionctrl,bai2025recammaster,yu2024viewcrafter,chen2025deepverse,mao2025yume}. World-modeling approaches represent dynamic scenes in unified 3D or 4D spaces for geometry-consistent camera and object motion~\citep{ren2025gen3c, zheng2026versecrafter}. These methods establish 4D representations as effective controls for synthesizing plausible camera and object motion. PREX extends this setting to faithful editing by assigning preservation and synthesis roles according to source-observation support.

\noindent\textbf{Motion- and Trajectory-Controlled Video Generation.} Motion-controlled video generation explores conditions such as bounding boxes, masks, optical flow, human poses, and user-specified trajectories. Point trajectories are attractive because they provide a flexible interface for sparse object motion and dense scene motion. Early methods rely on 2D strokes or tracks~\citep{yin2023dragnuwa,wang2024boximator,wu2024draganything,xing2025motioncanvas,li2025magicmotion,zhang2025motionpro,chu2025wan,wang2025ati,zhang2025flextraj}, while recent works introduce 3D-aware trajectories for improved depth reasoning, occlusion handling, and viewpoint control~\citep{gu2025diffusion,lee2026generative}. PREX complements these motion-control objectives with evidence-aware preservation, retaining valid source content and synthesizing regions unsupported after a 4D edit.

\noindent\textbf{Video Editing.} Video editing methods achieve results in appearance editing, object replacement, insertion, removal, and local deformation~\citep{geyer2023tokenflow,qi2023fatezero,yang2023rerender,wang2025videodirector,li2024vidtome,zhou2023propainter}. These image-plane editors emphasize semantic or local changes while retaining the source layout. Novel view synthesis and camera-controlled video-to-video methods render or generate videos from new viewpoints by warping source observations and inpainting disoccluded regions~\citep{mildenhall2022nerf,kerbl20233d,wu20244d,wu2024reconfusion,yu2024viewcrafter,bai2025recammaster,lin2026vista4d,yang2026neoverse}. PREX targets visibility changes induced by 4D camera and object edits, modeling source-supported, revealed, and out-of-view regions with region-aware geometric guidance and observation-backed appearance cues.

\begin{figure*}[t]
  \centering
  \includegraphics[width=0.9\linewidth]{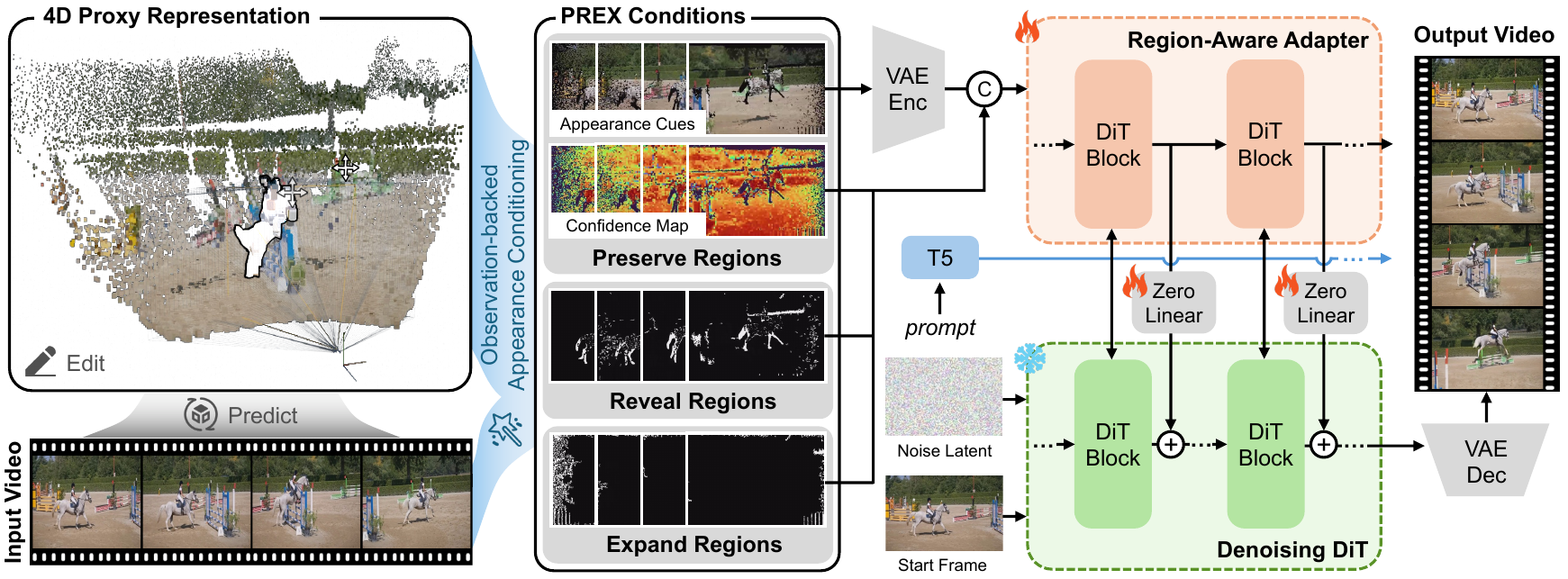}
  \caption{\textbf{An overview of PREX pipeline.} PREX conditions a frozen video diffusion model with seperated regions, observation-backed cues and confidence maps through a Region-aware Adapter.}
  \vspace{-0.95em}
  \label{fig:main}
\end{figure*}

\begin{figure*}[t]
  \centering

  \begin{minipage}[t]{0.71\linewidth}
    \centering
    \includegraphics[width=\linewidth]{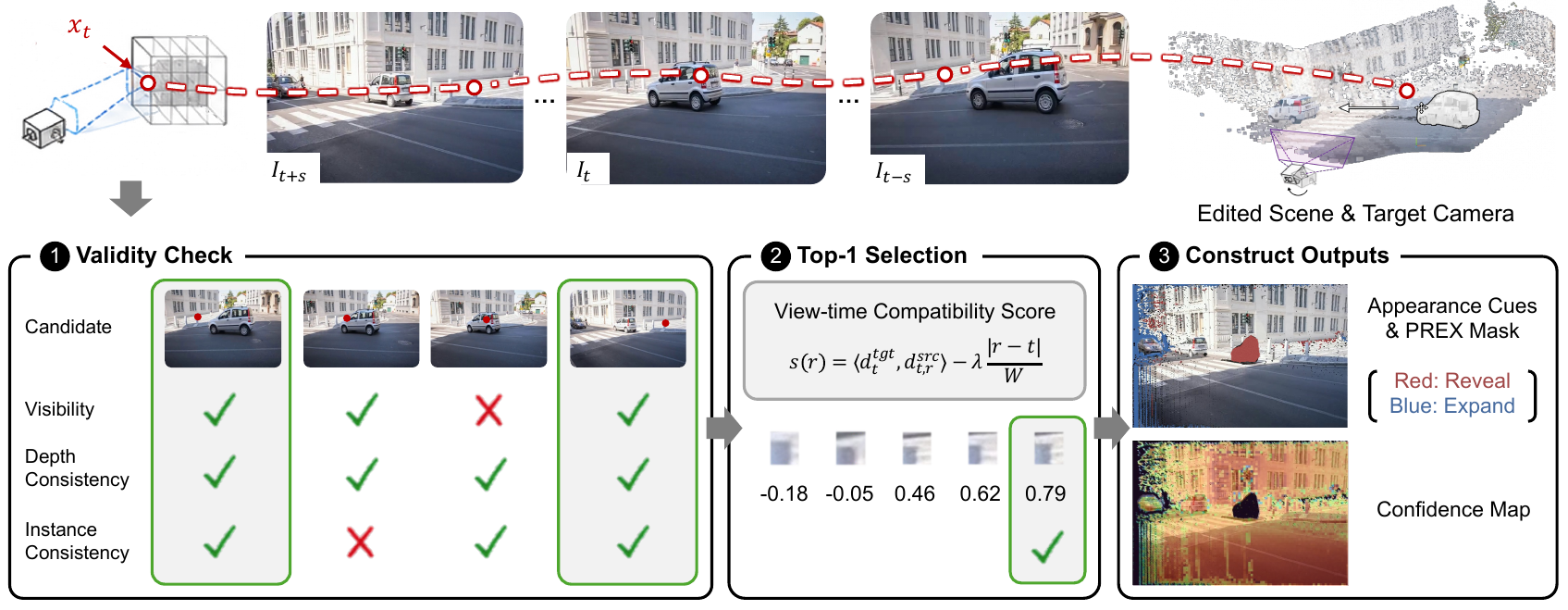}
    \caption{Design of \textbf{Observation-backed Appearance Rendering} for construction of geometric conditions.}
    \label{fig:obs}
  \end{minipage}%
  \hspace{0.02\linewidth}
  \begin{minipage}[t]{0.25\linewidth}
    \centering
    \includegraphics[width=\linewidth]{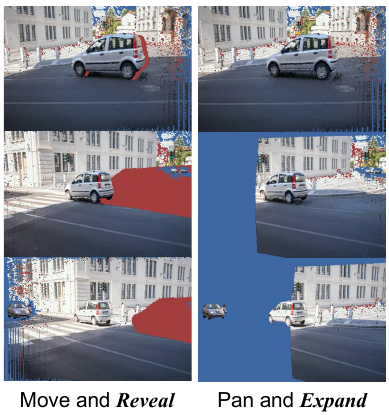}
    \caption{Proxy-task for supervised training.}
    \label{fig:train_detail}
  \end{minipage}

  \vspace{-1em}
\end{figure*}

\section{Method}

\label{sec:method_overview}

Given a source video $V=\{I_t\}_{t=1}^{T}$, PREX generates an edited video $\hat V$ that follows user-specified camera and object edits while preserving source appearance wherever valid visual evidence remains. Regions that become newly visible or extend beyond the observed scene are instead synthesized by a video diffusion model. To expose this distinction explicitly, PREX first reconstructs the source video into a static--dynamic 4D proxy $\mathcal S$ with aligned tracks $\mathcal T$, and applies the user edit $\mathcal E$ to obtain the edited proxy $\mathcal S'$. The overall pipeline is
\begin{equation} V \rightarrow (\mathcal S,\mathcal T) \xrightarrow{\mathcal E} \mathcal S' \rightarrow \{\mathcal R_t\}_{t=1}^{T} \xrightarrow{\mathcal A_\theta,\,\epsilon_{\mathrm{frozen}}} \hat V, \label{eq:pipeline_overview} \end{equation}
where $\mathcal A_\theta$ is a trainable adapter and $\epsilon_{\mathrm{frozen}}$ is a frozen video diffusion backbone. We instantiate the reconstruction stage with Uni4D~\citep{yao2025uni4d}; the downstream controls consume its standardized editable 4D proxy. Representation and implementation details are provided in Appendix~\ref{app:reconstruction}.

PREX makes two distinctions often conflated by a coarse 4D render. Geometric coverage does not by itself imply a target pixel has a valid source observation for appearance preservation, and unsupported pixels can lie within the scene extent or outside. PREX exposes both distinctions explicitly: observation-validity tests determine Preserve support, scene extent separates Reveal from Expand, and the roles are used consistently for conditioning, proxy-task construction, and evaluation.

The key interface between the edited 4D proxy and the generative model is an evidence-aware, frame-aligned control state. For each target camera $\Pi_t$, PREX projects $\mathcal S'$ into
\begin{equation} \mathcal R_t=\{C_t^{rgb},C_t^{conf},M_t^P,M_t^R,M_t^E\}. \label{eq:framewise_control_state} \end{equation}
where $C_t^{rgb}$ provides source-backed appearance cues and $C_t^{conf}$ encodes their reliability. The disjoint masks $M_t^P$, $M_t^R$, and $M_t^E$ partition the target frame into three evidence regimes: \emph{Preserve} regions supported by reliable source observations, \emph{Reveal} regions exposed by the edit but unsupported because of prior occlusion, and \emph{Expand} regions lying beyond the observed scene support. This representation allows the adapter to distinguish appearance that should be faithfully retained from content that must be synthesized, while leaving the pretrained video prior frozen. Figure~\ref{fig:main} summarizes the complete pipeline.

\subsection{Observation-backed Appearance and Evidence-aware Roles}
\label{sec:ob_appearance_short}

Coarse 4D renders can blur high-frequency appearance and propagate invalid colors into disoccluded regions. PREX therefore recovers appearance directly from valid source observations rather than treating the reconstruction as uniformly reliable. For each target pixel $x$, we trace its edited 4D point back to candidate source frames $r\in\mathcal W_t$; for edited dynamic instances, this correspondence additionally applies the inverse object transformation. A source observation is considered valid only if it passes image-domain visibility, depth-consistency, and instance-identity tests, denoted by $\nu_{t,r}(x)=1$. The Preserve mask is therefore defined by the existence of valid source evidence,
\begin{equation}
M_t^P(x)=\ind{\max_{r\in\mathcal W_t}\nu_{t,r}(x)=1},
\label{eq:preserve_mask}
\end{equation}
rather than simply the support of a rendered point cloud. Among valid observations, we select the source frame that balances viewpoint agreement and temporal proximity,
\begin{equation}
r^*(x)=\arg\max_{r:\nu_{t,r}(x)=1}
\left[
\left\langle d_t^{\mathrm{tgt}},d_{t,r}^{\mathrm{src}}\right\rangle
-\lambda |r-t|/W
\right],
\label{eq:view_time_score}
\end{equation}
and write the corresponding observed color into $C_t^{rgb}(x)$. Pixels without valid source observations are not treated as reliable appearance constraints and are instead left to the generative model. Projection tests and compositing details are provided in Appendix~\ref{app:observation_details}.

Binary validity alone does not capture the varying reliability of supported observations. PREX therefore assigns each preserved pixel a continuous confidence based on coverage $g_t^{cov}$, instance purity $g_t^{pur}$, local depth variation $g_t^{std}$, and source--target view agreement:
\begin{equation}
\begin{split}
C_t^{conf}(x)={}&M_t^P(x)\,g_t^{cov}(x)g_t^{pur}(x)
\exp\!\left(-g_t^{std}(x)/\tau\right)\\
&\cdot\left[\alpha+(1-\alpha)
\max\!\left(0,\langle\hat{\mathbf v}^{src},\hat{\mathbf v}^{tgt}\rangle\right)^\gamma\right].
\end{split}
\label{eq:calibrated_confidence}
\end{equation}
Thus, even within Preserve, observations with weak geometric support or large viewpoint mismatch can be down-weighted, while unsupported pixels receive zero confidence.

We further distinguish two types of unsupported regions. Let $M_t^S$ denote the support of the originally captured scene projected into the target view. PREX defines
\begin{equation}
M_t^R=(1-M_t^P)M_t^S,
\qquad
M_t^E=(1-M_t^P)(1-M_t^S).
\label{eq:reveal_expand_masks}
\end{equation}
Reveal pixels remain within the captured scene support but lack a valid observation, as occurs under disocclusion or object removal, whereas Expand pixels lie beyond that support and require out-of-view extrapolation. Together, $M_t^P$, $M_t^R$, and $M_t^E$ form a disjoint partition that tells the adapter both where source appearance can be trusted and what type of synthesis is required elsewhere. The construction of $M_t^S$ and additional confidence details are provided in Appendix~\ref{app:role_details}.

\subsection{Region-aware Diffusion Conditioning}

Given the evidence-aware controls $\{\mathcal R_t\}_{t=1}^{T}$, PREX converts them into soft latent conditions for a pretrained video diffusion model while keeping its VAE, text encoder, and denoising backbone frozen. We denote the temporally stacked controls by $C^{rgb}$, $C^{conf}$, $M^R$, and $M^E$. The observation-backed appearance $C^{rgb}$ is encoded by the frozen VAE, while confidence and region roles are embedded at the same latent resolution. Since the three roles form a partition, $M^P$ is redundant given $M^R$ and $M^E$, and the adapter input is
\begin{equation}
G=\mathrm{Concat}\!\left(
\mathcal E_{vae}(C^{rgb}),
\phi_{conf}(C^{conf}),
\phi_{mask}(M^R,M^E)
\right).
\label{eq:adapter_input}
\end{equation}
This representation separates appearance from its reliability and synthesis role: confidence specifies how strongly source-backed evidence should be trusted, whereas the region masks indicate whether unsupported content requires in-scene completion or out-of-view extrapolation.

The Region-Aware Adapter $\mathcal A_\theta$ maps $G$, diffusion timestep $n$, and text embedding $e_p$ to residual hints $h_\ell$ for selected backbone layers. For a frozen block $B_\ell$,
\begin{equation}
h_\ell=\mathcal A_\theta^\ell(G,n,e_p),\qquad
x_{\ell+1}=B_\ell(x_\ell,n,e_p)+h_\ell.
\label{eq:residual_injection}
\end{equation}

The output projections of the adapter are initialized near zero, so optimization starts close to the pretrained generator and progressively learns how to interpret PREX controls while keeping the pretrained backbone weights fixed. Importantly, these controls act as soft conditions rather than hard pixel constraints, allowing the model to retain reliable source appearance while synthesizing unsupported Reveal and Expand regions through the pretrained generative prior.

\subsection{Self-supervised Proxy Editing Tasks}
\label{sec}

To avoid collecting large-scale paired editing data, PREX derives supervision from training videos. Starting from the reconstructed proxy $\mathcal S$, we sample a role-oriented perturbation $q\in{P,R,E}$ that alters the available geometry, viewpoints, or observation support, and then build the control sequence $\tilde{\mathcal R}^{(q)}$ with the same procedure used at inference. Crucially, the perturbation changes only the evidence presented to the model, and the original video $V$ remains the reconstruction target. As shown in Figure~\ref{fig:train_detail}, these proxy tasks emulate the evidence regimes induced by edits. \emph{Preserve} retains near-original views and reliable observation support, encouraging faithful use of high-confidence appearance. \emph{Reveal} removes selected in-scene observations to simulate disocclusion and teaches completion without copying invalid foreground evidence. \emph{Expand} limits the available source views or temporal prefix while keeping held-out content as supervision, training the model to extrapolate beyond observed support. All three variants are interleaved within a curriculum, while geometry degradation and appearance jitter reduce dependence on reconstruction artifacts. Thus, PREX learns preservation, reveal completion, and expansion from ordinary videos without collecting paired editing supervision. Detailed proxy-task constructions are given in Appendix~\ref{app:proxy_tasks}.

\begin{table*}[t]
\centering
\caption{\textbf{Global visual quality on full PREBench test set.} VBench scores are reported.}
\label{tab:vbench}
\resizebox{\linewidth}{!}{
\begin{tabular}{lccccccc}
\toprule
\textbf{Method}
& \begin{tabular}{c}\textbf{Overall}$\uparrow$\\\textbf{Score}\end{tabular}
& \begin{tabular}{c}\textbf{Imaging}$\uparrow$\\\textbf{Quality}\end{tabular}
& \begin{tabular}{c}\textbf{Aesthetic}$\uparrow$\\\textbf{Quality}\end{tabular}
& \begin{tabular}{c}\textbf{Dynamic}$\uparrow$\\\textbf{Degree}\end{tabular}
& \begin{tabular}{c}\textbf{Motion}$\uparrow$\\\textbf{Smoothness}\end{tabular}
& \begin{tabular}{c}\textbf{Background}$\uparrow$\\\textbf{Consistency}\end{tabular}
& \begin{tabular}{c}\textbf{Subject}$\uparrow$\\\textbf{Consistency}\end{tabular} \\
\midrule
Perception-as-Control  & 68.96 & 54.90 & 43.15 & 49.49 & 96.83 & 86.44 & 82.94 \\
Yume                  & 70.97 & 58.80 & 42.06 & 50.51 & 95.51 & \underline{91.24} & 87.71 \\
DaS                   & 73.33 & 60.17 & 44.41 & 57.64 & 95.65 & \textbf{92.04} & \textbf{90.07} \\
GEN3C                 & \underline{74.68} & 53.95 & 44.53 & \underline{76.24} & 96.02 & 91.00 & 86.32 \\
VerseCrafter          & 73.72 & \textbf{61.49} & \underline{47.09} & 59.32 & \underline{96.98} & 90.35 & 87.10 \\
\midrule
\textbf{PREX (Ours)}                              & \textbf{77.99} & \underline{60.18} & \textbf{48.52} & \textbf{83.06} & \textbf{97.21} & 91.20 & \underline{87.82} \\
\bottomrule
\end{tabular}
}
\end{table*}

\begin{table*}[t]
\centering
\caption{\textbf{Region-aware evaluation on PREBench test set.}}
\label{tab:prebench}
\resizebox{\textwidth}{!}{
\begin{tabular}{lcccccccccc}
\toprule
\multirow{2}{*}{Method}
& \multicolumn{4}{c}{Preserve}
& \multicolumn{2}{c}{Reveal}
& \multicolumn{3}{c}{Expand}
& \multirow[c]{2}{*}[-0.65ex]{\shortstack[c]{PREBench\\Score $\uparrow$}} \\
\cmidrule(lr){2-5}
\cmidrule(lr){6-7}
\cmidrule(lr){8-10}
& P-LPIPS $\downarrow$
& P-DISTS $\downarrow$
& P-TempDrift $\downarrow$
& P-Dyn-LPIPS $\downarrow$
& R-Ghost $\downarrow$
& R-Seam $\downarrow$
& E-Temp $\downarrow$
& E-Seam $\downarrow$
& E-Copy $\downarrow$ & \\
\midrule
\multicolumn{11}{l}{\textit{Camera-only Set}} \\
NeoVerse & \textbf{0.1897} & \textbf{0.1172} & \underline{0.0572} & \textbf{0.0285} & 0.1625 & 0.0423 & 0.1282 & 0.0432 & \textbf{0.3629} & 56.61 \\
Vista4D & 0.3253 & 0.1526 & 0.1020 & 0.0362 & \textbf{0.1326} & \underline{0.0278} & \underline{0.0721} & \textbf{0.0282} & 0.4108 & \underline{57.21} \\
\textbf{PREX (Ours)} & \underline{0.2045} & \underline{0.1232} & \textbf{0.0524} & \underline{0.0302} & \underline{0.1388} & \textbf{0.0265} & \textbf{0.0702} & \underline{0.0342} & \underline{0.3845} & \textbf{60.98} \\
\midrule
\multicolumn{11}{l}{\textit{Full Set}} \\
Yume & 0.3768 & 0.1747 & 0.0816 & 0.0584 & 0.1629 & 0.0348 & 0.0871 & 0.0378 & 0.4128 & 53.42 \\
Perception-as-Control & 0.3915 & 0.1764 & 0.0827 & 0.0739 & 0.1746 & 0.0392 & 0.1017 & 0.0426 & 0.4285 & 51.20 \\
DaS & 0.4282 & 0.1881 & 0.0894 & 0.0919 & 0.1509 & \textbf{0.0214} & 0.1318 & 0.0342 & 0.4057 & 53.20 \\
GEN3C & \underline{0.3573} & \underline{0.1659} & \underline{0.0809} & \underline{0.0522} & 0.1931 & 0.0331 & \underline{0.0746} & \textbf{0.0273} & \underline{0.3931} & \underline{54.59} \\
VerseCrafter & 0.4757 & 0.2063 & 0.0953 & 0.1045 & \underline{0.1436} & 0.0326 & 0.0905 & 0.0356 & 0.4416 & 51.57 \\
\textbf{PREX (Ours)} & \textbf{0.2137} & \textbf{0.1199} & \textbf{0.0532} & \textbf{0.0319} & \textbf{0.1374} & \underline{0.0277} & \textbf{0.0638} & \underline{0.0308} & \textbf{0.3764} & \textbf{61.45} \\
\bottomrule
\end{tabular}
}
\vspace{-1em}
\end{table*}

\section{PREBench}

PREBench is a region-aware diagnostic benchmark that evaluates edit-induced regions according to their evidence roles, enabling targeted diagnosis of preservation drift, ghost leakage, boundary copying, and out-of-view temporal instability. It complements global perceptual evaluation by localizing these failures within Preserve, Reveal, and Expand regions.

For each editing case, we provide a source video, an edited 4D proxy, target cameras, and region masks that decompose the target video into \textit{Preserve}, \textit{Reveal}, and \textit{Expand} regions to evaluate whether a method preserves valid source evidence, completes newly revealed regions, and extrapolates out-of-view content in a temporally coherent manner. All region masks were first generated algorithmically and then manually reviewed and refined manually to ensure accurate region assignments. The resulting verified masks are fixed for each case and used consistently to evaluate all methods. The precise definitions of the benchmark metrics are provided in Appendix~\ref{app:prebench_metrics}.

\noindent\textbf{Preserve-region fidelity.}
Preserve regions contain pixels that remain supported by valid source observations after editing, and should retain the appearance, structure, and temporal behavior of the source video. 
We compare the generated video with source-backed reference observations within the Preserve mask. 
\textbf{P-LPIPS} and \textbf{P-DISTS} measure perceptual appearance drift, while \textbf{P-TempDrift} compares temporal residuals between the generated video and the reference to penalize inconsistent temporal drift. 
For edited dynamic objects that should remain visually preserved, we further report \textbf{P-Dyn-LPIPS} on dynamic preserve masks.

\noindent\textbf{Reveal-region completion.}
Reveal regions are unsupported pixels inside the original scene extent, such as areas exposed by object removal, relocation, or disocclusion. 
They should be completed from surrounding spatial and temporal context without propagating invalid source evidence. 
We report \textbf{R-Ghost} to quantify RGB similarity to ghost references from invalid coarse renderings or removed-object evidence, providing a diagnostic of residual ghosting in Reveal regions. 
We also report \textbf{R-Seam}, which measures discontinuities along the boundary between Reveal and Preserve regions, reflecting whether completion blends naturally with preserved content.

\noindent\textbf{Expand-region extrapolation.}
Expand regions appear when the target camera reveals areas outside the original field of view, requiring coherent scene extrapolation rather than in-scene completion.
We evaluate three failure modes.
\textbf{E-Temp} measures occlusion-aware, flow-compensated RGB and feature inconsistency over temporal correspondences in Expand regions.
\textbf{E-Seam} measures excess feature discontinuity across Expand--Preserve interfaces relative to the local variation on either side.
\textbf{E-Copy} measures signatures of unreliable evidence reuse by combining similarity to invalid RGB hints with texture-level similarity to the source-backed field-of-view boundary.

\vspace{-0.5em}

\section{Experiments}
\label{sec:experiments}

\noindent\textbf{Training and Evaluation Data.}
We train PREX on 10,000 training pairs sampled from seven public video datasets: 4,589 pairs from DynPose-100K, 4,590 from SpatialVID-HQ, and 821 from UVO, PointOdyssey, Dynamic Replica, DAVIS, and Spring. Source videos are split before pair construction, so all clips and editing variants derived from a source video remain in a single split. We evaluate all methods on the \textbf{PREBench} test set, which contains 350 editing cases drawn from real and synthetic videos. 
Among them, 150 cases involve camera-only editing, such as camera pan, zoom, orbit-like reshooting, and dolly/truck reshooting. 
The remaining 200 cases involve joint camera and object motion control, including object trajectory changes, scale or depth movement, object removal, and dynamicity switches.

\noindent\textbf{Implementation Details.}
PREX is built upon Wan2.1-I2V-14B video diffusion backbone, and we train only the proposed Region-Aware Adapter while keeping the backbone frozen. We train the adapter in two stages: 50 epochs of reconstruction training with a learning rate of $2 \times 10^{-5}$, followed by 50 epochs of proxy-task curriculum training with a learning rate of $5 \times 10^{-6}$. Training is conducted on 16 NVIDIA H200 GPUs and takes approximately 100 hours. All compared methods are evaluated using the same input resolution of $720 \times 480$ and test cases for fair comparison.

\begin{table}[t]
\centering

\begin{minipage}[t]{0.68\linewidth}
\vspace{0pt}
\centering
\captionof{table}{\textbf{4D control quality.} Including Camera-only and Joint Camera/Object control.}
\label{tab:control}
\scriptsize
\setlength{\tabcolsep}{2.5pt}
\renewcommand{\arraystretch}{0.90}
\resizebox{\linewidth}{!}{
\begin{tabular}{llccc}
\toprule
Setting & Method
& Cam-RotErr $\downarrow$
& Cam-TransErr $\downarrow$
& ObjMC $\downarrow$ \\
\midrule
\multirow{7}{*}{Camera-only Set}
& DaS                  & 7.2936 & 6.0120 & -- \\
& GEN3C                & 2.1767 & 2.9738 & -- \\
& VerseCrafter         & 7.1165 & 6.8450 & -- \\
& FlashWorld           & 6.3761 & 5.2798 & -- \\
& Voyager              & 3.7782 & 5.0294 & -- \\
& NeoVerse             & \underline{1.4736} & \textbf{1.8353} & -- \\
& \textbf{PREX (Ours)} & \textbf{1.2345} & \underline{2.7236} & -- \\
\midrule
\multirow{4}{*}{\shortstack{Joint \\ Camera/Object Set}}
& DaS                  & 8.7452 & 5.8361 & 3.4257 \\
& GEN3C                & \underline{2.5236} & \underline{2.8738} & \underline{2.6844} \\
& VerseCrafter         & 6.9662 & 4.7655 & 2.9745 \\
& \textbf{PREX (Ours)} & \textbf{1.5676} & \textbf{1.5986} & \textbf{2.5923} \\
\bottomrule
\end{tabular}
}
\end{minipage}
\hfill
\begin{minipage}[t]{0.29\linewidth}
\vspace{0pt}
\centering
\captionof{figure}{\textbf{PREX win rates in 2AFC} (\%).}
\label{fig:usr_study}
\begin{tikzpicture}[x=0.38mm,y=0.35cm,font=\tiny]
  \def\prefrow#1#2#3#4{%
    \node[anchor=east] at (28,#1) {#2};
    \fill[black!8] (32,{#1-0.25}) rectangle (82,{#1+0.25});
    \fill[#4] (32,{#1-0.25}) rectangle ({32+(#3-50)},{#1+0.25});
    \node[anchor=west] at (84,#1) {#3};
  }

  \node[anchor=west,font=\scriptsize\bfseries,text=motionBlue] at (0,11.2) {Motion alignment};
  \prefrow{10.2}{DaS}{82.3}{motionBlue}
  \prefrow{9.35}{GEN3C}{87.5}{motionBlue}
  \prefrow{8.5}{VerseCrafter}{90.1}{motionBlue}

  \node[anchor=west,font=\scriptsize\bfseries,text=contextGreen] at (0,7.35) {Context preservation};
  \prefrow{6.35}{DaS}{81.5}{contextGreen}
  \prefrow{5.5}{GEN3C}{74.8}{contextGreen}
  \prefrow{4.65}{VerseCrafter}{67.8}{contextGreen}

  \node[anchor=west,font=\scriptsize\bfseries,text=qualityOrange] at (0,3.5) {Visual quality};
  \prefrow{2.5}{DaS}{84.0}{qualityOrange}
  \prefrow{1.65}{GEN3C}{83.3}{qualityOrange}
  \prefrow{0.8}{VerseCrafter}{87.6}{qualityOrange}

  \draw[black!35] (32,0.15) -- (82,0.15);
  \node[anchor=north,text=black!60] at (32,0.05) {50};
  \node[anchor=north,text=black!60] at (82,0.05) {100};
\end{tikzpicture}
\end{minipage}

\vspace{-0.5em}
\end{table}

\begin{table*}[t]
\centering
\caption{\textbf{Ablation study of main components on full PREBench test set.}}
\label{tab:ablation}
\resizebox{\textwidth}{!}{
\begin{tabular}{lccccccccccc}
\toprule
\multirow{2}{*}{Variant}
& \multirow{2}{*}{VBench $\uparrow$}
& \multicolumn{4}{c}{Preserve}
& \multicolumn{2}{c}{Reveal}
& \multicolumn{3}{c}{Expand}
& \multirow[c]{2}{*}[-0.65ex]{\shortstack[c]{PREBench\\Score $\uparrow$}} \\
\cmidrule(lr){3-6}
\cmidrule(lr){7-8}
\cmidrule(lr){9-11}
& 
& P-LPIPS $\downarrow$
& P-DISTS $\downarrow$
& P-TempDrift $\downarrow$
& P-Dyn-LPIPS $\downarrow$
& R-Ghost $\downarrow$
& R-Seam $\downarrow$
& E-Temp $\downarrow$
& E-Seam $\downarrow$
& E-Copy $\downarrow$ & \\
\midrule
Plain Wan2.1-ControlNet         & 75.27 & 0.3925 & 0.1682 & 0.0872 & 0.0840 & 0.1889 & 0.0437 & 0.1321 & 0.0508 & 0.3926 & 49.39 \\
w/o adapter design            & 74.21 & 0.3416 & 0.1528 & 0.0744 & 0.0625 & 0.1723 & 0.0342 & 0.0789 & 0.0475 & 0.3887 & 53.98 \\
w/o o-b appearance cues       & 76.90 & 0.3821 & 0.1765 & 0.0811 & 0.0576 & \underline{0.1382} & 0.0320 & \textbf{0.0636} & \underline{0.0341} & \textbf{0.3721} & 56.29 \\
w/o proxy-task training     & 75.75 & 0.2152 & \underline{0.1205} & 0.0565 & 0.0322 & 0.1688 & 0.0355 & 0.1077 & 0.0427 & 0.3843 & 56.80 \\
w/ binary role regions           & \underline{77.89} & 0.2144 & 0.1242 & \underline{0.0534} & \textbf{0.0319} & 0.1632 & \underline{0.0283} & 0.0692 & 0.0367 & 0.3883 & \underline{59.54} \\
w/o role regions              & 77.62 & \underline{0.2140} & 0.1276 & 0.0544 & \underline{0.0320} & 0.1598 & 0.0299 & 0.0708 & 0.0385 & 0.3804 & 59.26 \\
w/o confidence                & 77.27 & 0.2141 & 0.1218 & 0.0589 & 0.0339 & 0.1433 & 0.0292 & 0.0841 & 0.0368 & 0.3768 & 59.34 \\
\midrule
Full PREX                     & \textbf{77.99} & \textbf{0.2137} & \textbf{0.1199} & \textbf{0.0532} & \textbf{0.0319} & \textbf{0.1374} & \textbf{0.0277} & \underline{0.0638} & \textbf{0.0308} & \underline{0.3764} & \textbf{61.45} \\
\bottomrule
\end{tabular}
}

\vspace{-0.5em}
\end{table*}

\noindent\textbf{Baselines.}
We compare with Perception-as-Control~\citep{chen2025perception}, Yume~\citep{mao2025yume}, DaS~\citep{gu2025diffusion}, GEN3C~\citep{ren2025gen3c}, and VerseCrafter~\citep{zheng2026versecrafter}. For camera-only control, we additionally include FlashWorld~\citep{li2025flashworld}, Voyager~\citep{huang2025voyager}, NeoVerse~\citep{yang2026neoverse}, and Vista4D~\citep{lin2026vista4d}.

\begin{figure*}[t]
  \centering
  \includegraphics[width=\textwidth]{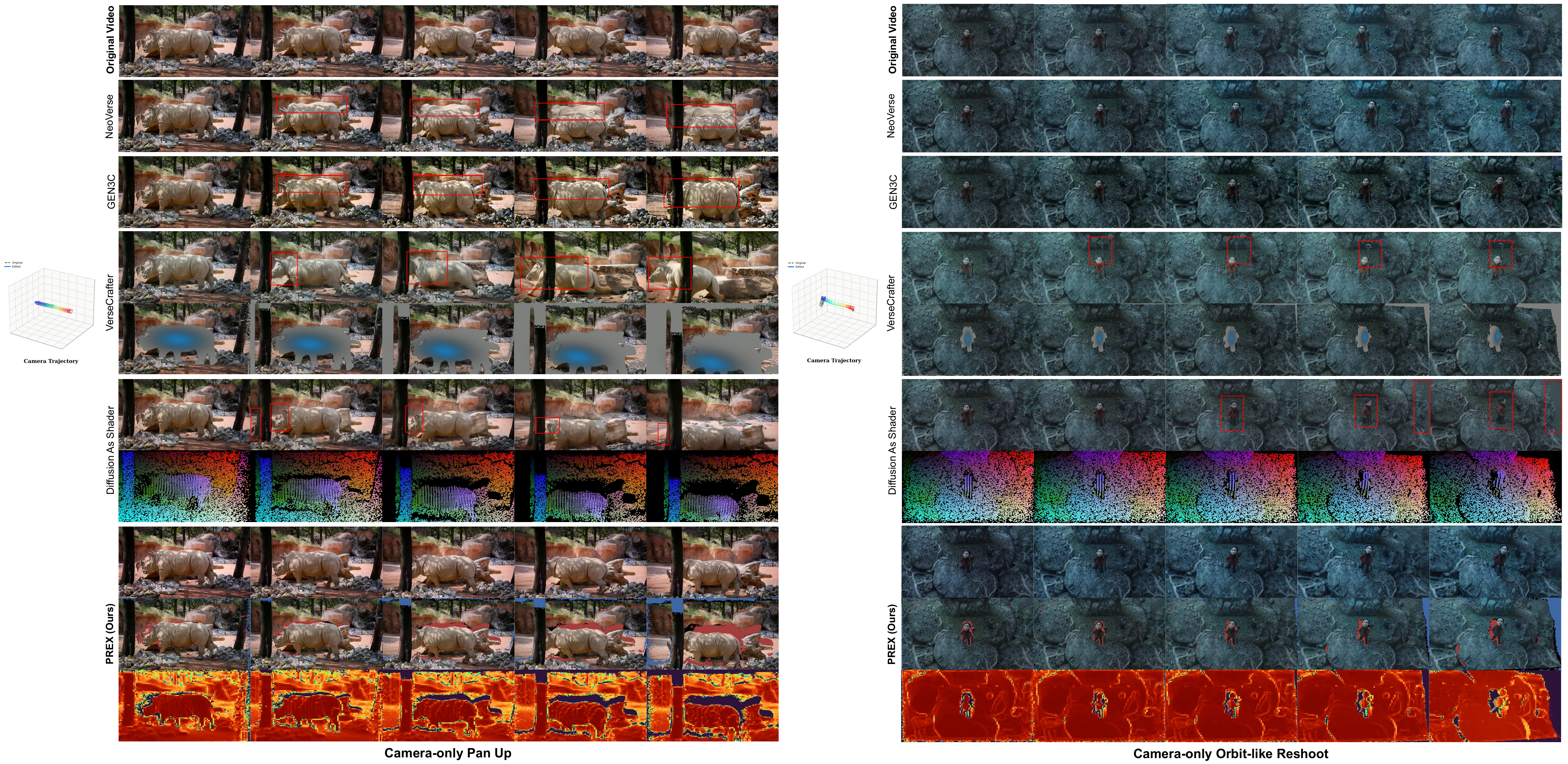}
  \caption{\textbf{Qualitative comparison of camera-only motion control on PREBench.}}
  \label{fig:qualitative_main1}
  \vspace{-0.2em}
\end{figure*}

\begin{figure*}[t]
  \centering
\includegraphics[width=\textwidth]{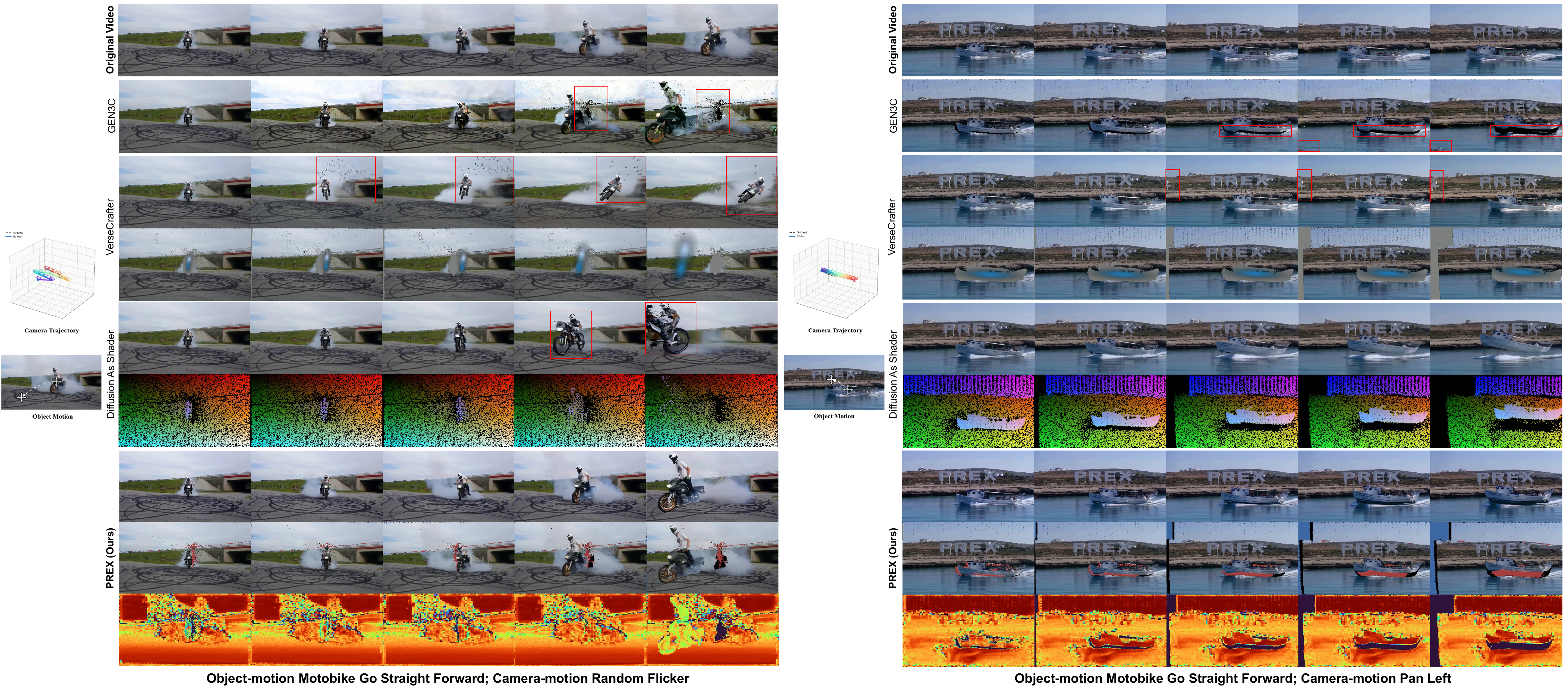}
  \caption{\textbf{Qualitative comparison of camera-object joint motion control on PREBench.}}
  \label{fig:qualitative_main2}
  \vspace{-1em}
\end{figure*}

\subsection{Main Results}

\noindent\textbf{Global Video Quality} Global metrics complement region-specific diagnostics by measuring perceptual realism and temporal smoothness. As shown in Table~\ref{tab:vbench}, PREX achieves the highest overall VBench score among the evaluated methods.

\noindent\textbf{Region-aware Evaluation on PREBench} We further evaluate region-specific editing fidelity using PREBench. As shown in Table~\ref{tab:prebench}, in the camera-only setting, PREX obtains the highest PREBench Score of 60.98, outperforming Vista4D (57.21) and NeoVerse (56.61). PREX ranks first on P-TempDrift, R-Seam, and E-Temp, and second on the remaining six component metrics, yielding the strongest region-balanced performance.

On the full PREBench set, PREX achieves the best performance on all Preserve-region metrics, the lowest R-Ghost, E-Temp, and E-Copy, and the second-best R-Seam and E-Seam. Its Score reaches 61.45, far exceeding top comparator GEN3C (54.59), demonstrating more reliable preservation and unsupported-region synthesis across camera-only and joint camera/object edits.


\noindent\textbf{User Study} We conduct a human perceptual evaluation with 40 subjects using 2AFC to assess 20 PREBench motion editing cases, including camera and object motion editing. Subjects assessed output quality based on three aspects: (i) alignment with desired motion, (ii) preservation of input context, and (iii) perceived visual quality. As reported in Fig.~\ref{fig:usr_study}, our method shows a higher preference over the baselines across all three key aspects of the video editing task.

\noindent\textbf{Camera and Object Motion Control} \label{sec:control_results} Following VerseCrafter~\citep{zheng2026versecrafter}, we further evaluate 4D control quality using the same baselines. As shown in Table~\ref{tab:control}, PREX achieves the lowest camera rotation error in the camera-only setting and the best camera rotation, camera translation, and object motion control scores in the joint setting. 
These results indicate that PREX can better follow target 4D edits while maintaining accurate camera and object motion control.

\begin{figure*}[t!]

  \begin{minipage}[t]{0.48\linewidth}
    \centering
    \includegraphics[width=\linewidth]{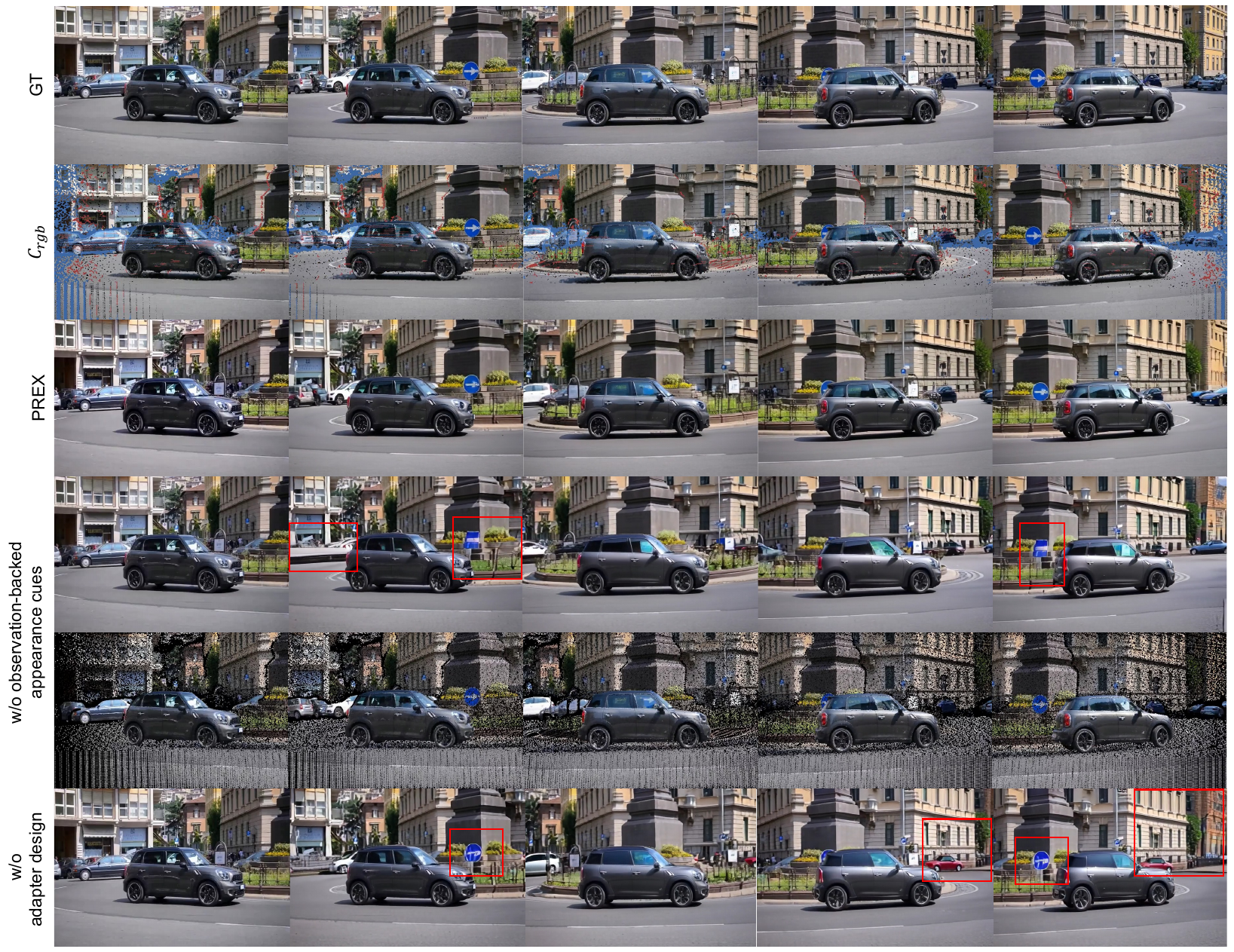}
    \captionof{figure}{\textbf{Qualitative ablation of observation-backed appearance cues and adapter design.}}
    \label{fig:ablation1}
  \end{minipage}%
  \hfill%
  \begin{minipage}[t]{0.48\linewidth}
    \centering
    \includegraphics[width=\linewidth]{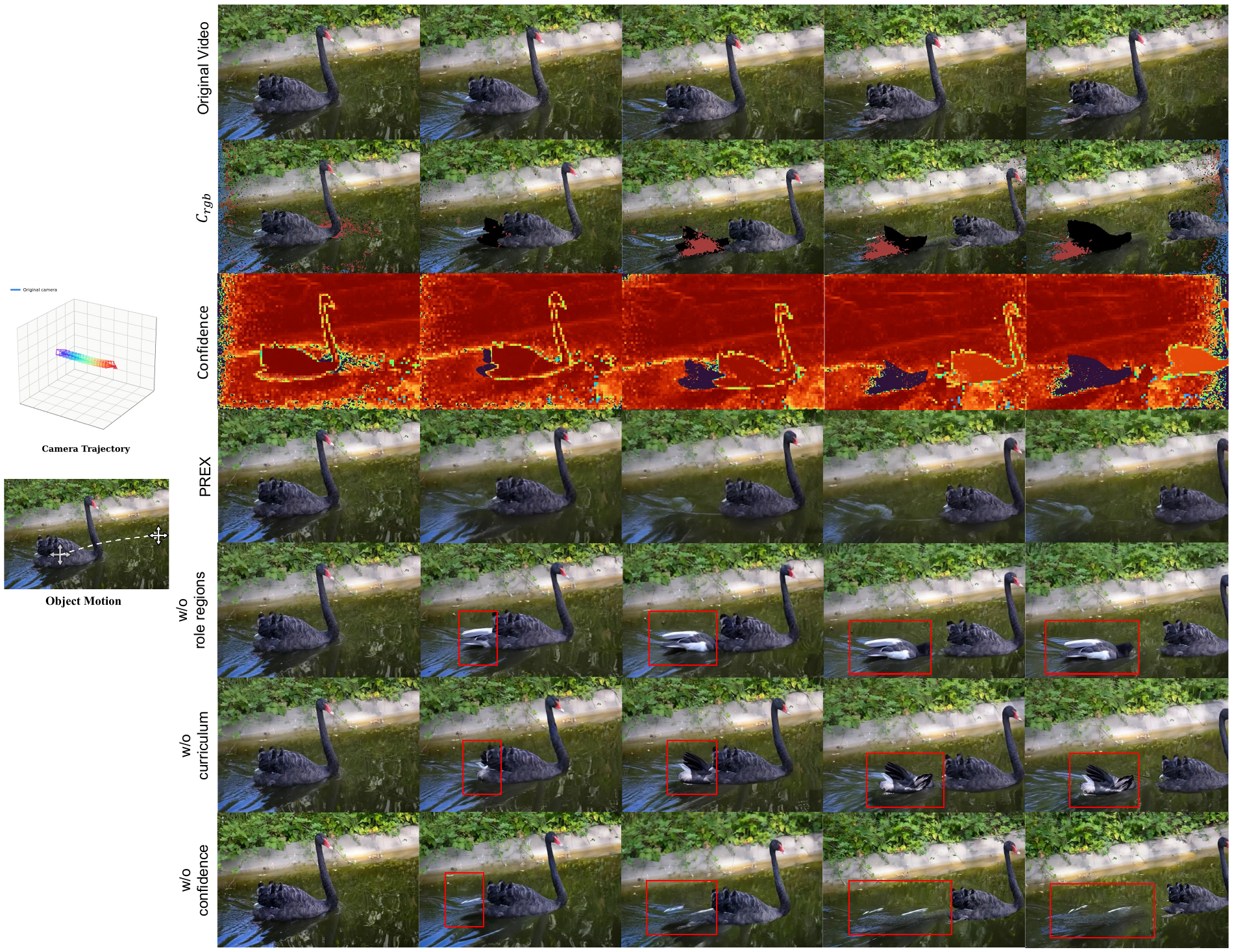}
    \captionof{figure}{\textbf{Qualitative ablation of role regions, proxy-task, and confidence maps.}}
    \label{fig:ablation2}
  \end{minipage}

  \vspace{-1em}
\end{figure*}

\noindent\textbf{Qualitative Comparisons} \label{sec:qualitative} Figures~\ref{fig:qualitative_main1}, ~\ref{fig:qualitative_main2} show qualitative comparisons on representative PREBench cases. We visualize the input video, target 4D control, region masks, and generated outputs. Compared with baselines, PREX better preserves source-backed regions, suppresses ghost artifacts in Reveal regions, and produces temporally coherent expansion outside the original field of view.

\subsection{Ablation Study}
\label{sec:ablation}

As shown in Table~\ref{tab:ablation}, we compare PREX with a matched Plain Wan2.1-ControlNet using the same backbone, first-frame condition, data, 4D proxy, and budget; the plain baseline directly conditions on the raw RGB render. For the component variants, “w/o adapter design” replaces the Region-Aware Adapter with a standard ControlNet, “w/o role regions” and “w/o confidence” remove the corresponding channels and retrain, and “w/o proxy-task training” uses only reconstruction pairs with all other settings unchanged. Full PREX improves VBench from 75.27 to 77.99, raises PREBench Score from 49.39 to 61.45, and reduces all region-aware errors, including P-LPIPS by 45.6\%, R-Ghost by 27.3\%, E-Temp by 51.7\%, and E-Seam by 39.4\%. The component ablations show complementary contributions: removing appearance cues increases P-LPIPS from 0.2137 to 0.3821 and P-Dyn-LPIPS from 0.0319 to 0.0576. Removing role regions raises R-Ghost and E-Seam by 16.3\% and 25.0\%, respectively. Removing confidence raises E-Temp by 31.8\%, while removing proxy-task training raises the same metric by 68.8\%. Without the adapter design, VBench drops to 74.21 and P-LPIPS rises to 0.3416. “w/ binary role regions” retains Preserve while merging Reveal and Expand into one unsupported region largely maintains preservation fidelity but increases R-Ghost and E-Seam errors by 18.8\% and 19.2\%, supporting distinct Reveal and Expand conditioning.

The qualitative ablations in Figures~\ref{fig:ablation1} and~\ref{fig:ablation2} show how the components affect different evidence regimes. In the driving scene, removing observation-backed appearance cues leads to a loss of fine details in the source-visible street scene, while replacing the adapter produces localized distortions around background objects and vehicle edges. In the swan edit, removing role regions leaves a duplicate in the newly revealed water; without proxy-task training, a fragmented residual remains, and without confidence, a faint copy persists. Full PREX preserves the observed scene while completing the area vacated by the moving swan, consistent with the component trends in Table~\ref{tab:ablation}.

\section{Conclusion}

We presented PREX, a region-aware framework for faithful 4D video editing. PREX addresses Evidence-Role Mismatch by separating target videos into Preserve, Reveal, and Expand regions, and combines observation-backed cues, geometric confidence, and a region-aware adapter within a frozen video diffusion backbone. We also introduced PREBench, a diagnostic benchmark with region-aware metrics. Experiments show that PREX reduces preservation drift, ghost leakage, and expansion artifacts while achieving the highest overall VBench score among the evaluated methods, demonstrating the value of disentangling evidence reliability from editing role.

\noindent\textbf{Limitations and Future Work.} PREX remains limited by upstream 4D errors and challenging occlusions, non-rigid motion, and transparent or reflective objects. Promising directions include improving robustness to unreliable 4D evidence and using PREBench as a reward signal for reinforcement-learning post-training of 4D-driven video diffusion models.

\clearpage

\bibliography{references}

@article{he2024cameractrl,
  title={Cameractrl: Enabling camera control for text-to-video generation},
  author={He, Hao and Xu, Yinghao and Guo, Yuwei and Wetzstein, Gordon and Dai, Bo and Li, Hongsheng and Yang, Ceyuan},
  journal={arXiv preprint arXiv:2404.02101},
  year={2024}
}

@inproceedings{wang2024motionctrl,
  title={Motionctrl: A unified and flexible motion controller for video generation},
  author={Wang, Zhouxia and Yuan, Ziyang and Wang, Xintao and Li, Yaowei and Chen, Tianshui and Xia, Menghan and Luo, Ping and Shan, Ying},
  booktitle={ACM SIGGRAPH 2024 Conference Papers},
  pages={1--11},
  year={2024}
}

@inproceedings{bai2025recammaster,
  title={Recammaster: Camera-controlled generative rendering from a single video},
  author={Bai, Jianhong and Xia, Menghan and Fu, Xiao and Wang, Xintao and Mu, Lianrui and Cao, Jinwen and Liu, Zuozhu and Hu, Haoji and Bai, Xiang and Wan, Pengfei and others},
  booktitle={Proceedings of the IEEE/CVF International Conference on Computer Vision},
  pages={14834--14844},
  year={2025}
}

@article{yu2024viewcrafter,
  title={Viewcrafter: Taming video diffusion models for high-fidelity novel view synthesis},
  author={Yu, Wangbo and Xing, Jinbo and Yuan, Li and Hu, Wenbo and Li, Xiaoyu and Huang, Zhipeng and Gao, Xiangjun and Wong, Tien-Tsin and Shan, Ying and Tian, Yonghong},
  journal={arXiv preprint arXiv:2409.02048},
  year={2024}
}

@article{chen2025deepverse,
  title={Deepverse: 4d autoregressive video generation as a world model},
  author={Chen, Junyi and Zhu, Haoyi and He, Xianglong and Wang, Yifan and Zhou, Jianjun and Chang, Wenzheng and Zhou, Yang and Li, Zizun and Fu, Zhoujie and Pang, Jiangmiao and others},
  journal={arXiv preprint arXiv:2506.01103},
  year={2025}
}

@article{mao2025yume,
  title={Yume: An interactive world generation model},
  author={Mao, Xiaofeng and Lin, Shaoheng and Li, Zhen and Li, Chuanhao and Peng, Wenshuo and He, Tong and Pang, Jiangmiao and Chi, Mingmin and Qiao, Yu and Zhang, Kaipeng},
  journal={arXiv preprint arXiv:2507.17744},
  year={2025}
}

@article{zheng2026versecrafter,
  title={VerseCrafter: Dynamic Realistic Video World Model with 4D Geometric Control},
  author={Zheng, Sixiao and Yin, Minghao and Hu, Wenbo and Li, Xiaoyu and Shan, Ying and Fu, Yanwei},
  journal={arXiv preprint arXiv:2601.05138},
  year={2026}
}

@inproceedings{ren2025gen3c,
  title={Gen3c: 3d-informed world-consistent video generation with precise camera control},
  author={Ren, Xuanchi and Shen, Tianchang and Huang, Jiahui and Ling, Huan and Lu, Yifan and Nimier-David, Merlin and M{\"u}ller, Thomas and Keller, Alexander and Fidler, Sanja and Gao, Jun},
  booktitle={Proceedings of the IEEE/CVF Conference on Computer Vision and Pattern Recognition},
  pages={6121--6132},
  year={2025}
}

@article{wang2024boximator,
  title={Boximator: Generating rich and controllable motions for video synthesis},
  author={Wang, Jiawei and Zhang, Yuchen and Zou, Jiaxin and Zeng, Yan and Wei, Guoqiang and Yuan, Liping and Li, Hang},
  journal={arXiv preprint arXiv:2402.01566},
  year={2024}
}

@inproceedings{wu2024draganything,
  title={Draganything: Motion control for anything using entity representation},
  author={Wu, Weijia and Li, Zhuang and Gu, Yuchao and Zhao, Rui and He, Yefei and Zhang, David Junhao and Shou, Mike Zheng and Li, Yan and Gao, Tingting and Zhang, Di},
  booktitle={European Conference on Computer Vision},
  pages={331--348},
  year={2024},
  organization={Springer}
}

@inproceedings{xing2025motioncanvas,
  title={Motioncanvas: Cinematic shot design with controllable image-to-video generation},
  author={Xing, Jinbo and Mai, Long and Ham, Cusuh and Huang, Jiahui and Mahapatra, Aniruddha and Fu, Chi-Wing and Wong, Tien-Tsin and Liu, Feng},
  booktitle={Proceedings of the Special Interest Group on Computer Graphics and Interactive Techniques Conference Conference Papers},
  pages={1--11},
  year={2025}
}

@inproceedings{li2025magicmotion,
  title={Magicmotion: Controllable video generation with dense-to-sparse trajectory guidance},
  author={Li, Quanhao and Xing, Zhen and Wang, Rui and Zhang, Hui and Dai, Qi and Wu, Zuxuan},
  booktitle={Proceedings of the IEEE/CVF International Conference on Computer Vision},
  pages={12112--12123},
  year={2025}
}

@inproceedings{zhang2025motionpro,
  title={Motionpro: A precise motion controller for image-to-video generation},
  author={Zhang, Zhongwei and Long, Fuchen and Qiu, Zhaofan and Pan, Yingwei and Liu, Wu and Yao, Ting and Mei, Tao},
  booktitle={Proceedings of the Computer Vision and Pattern Recognition Conference},
  pages={27957--27967},
  year={2025}
}

@inproceedings{chu2025wan,
  title={Wan-Move: Motion-Controllable Video Generation via Latent Trajectory Guidance},
  author={Chu, Ruihang and He, Yefei and Chen, Zhekai and Zhang, Shiwei and Xu, Xiaogang and Xia, Bin and Wang, Dingdong and Yi, Hongwei and Liu, Xihui and Zhao, Hengshuang and others},
  booktitle={Advances in Neural Information Processing Systems},
  year={2025}
}

@inproceedings{jiang2025vace,
  title={VACE: All-in-One Video Creation and Editing},
  author={Jiang, Zeyinzi and Han, Zhen and Mao, Chaojie and Zhang, Jingfeng and Pan, Yulin and Liu, Yu},
  booktitle={Proceedings of the IEEE/CVF International Conference on Computer Vision},
  pages={17191--17202},
  year={2025}
}

@article{yin2023dragnuwa,
  title={Dragnuwa: Fine-grained control in video generation by integrating text, image, and trajectory},
  author={Yin, Shengming and Wu, Chenfei and Liang, Jian and Shi, Jie and Li, Houqiang and Ming, Gong and Duan, Nan},
  journal={arXiv preprint arXiv:2308.08089},
  year={2023}
}

@article{zhang2025flextraj,
  title={FlexTraj: Image-to-Video Generation with Flexible Point Trajectory Control},
  author={Zhang, Zhiyuan and Wang, Can and Chen, Dongdong and Liao, Jing},
  journal={arXiv preprint arXiv:2510.08527},
  year={2025}
}

@inproceedings{lee2026generative,
  title={Generative Video Motion Editing with 3D Point Tracks},
  author={Lee, Yao-Chih and Zhang, Zhoutong and Huang, Jiahui and Wang, Jui-Hsien and Lee, Joon-Young and Huang, Jia-Bin and Shechtman, Eli and Li, Zhengqi},
  booktitle={Proceedings of the IEEE/CVF Conference on Computer Vision and Pattern Recognition},
  pages={18306--18318},
  year={2026}
}

@inproceedings{burgert2026motionv2v,
  title={MotionV2V: Editing Motion in a Video},
  author={Burgert, Ryan and Herrmann, Charles and Cole, Forrester and Ryoo, Michael S. and Wadhwa, Neal and Voynov, Andrey and Ruiz, Nataniel},
  booktitle={Proceedings of the IEEE/CVF Conference on Computer Vision and Pattern Recognition},
  year={2026}
}

@inproceedings{gu2025diffusion,
  title={Diffusion as shader: 3d-aware video diffusion for versatile video generation control},
  author={Gu, Zekai and Yan, Rui and Lu, Jiahao and Li, Peng and Dou, Zhiyang and Si, Chenyang and Dong, Zhen and Liu, Qifeng and Lin, Cheng and Liu, Ziwei and others},
  booktitle={Proceedings of the Special Interest Group on Computer Graphics and Interactive Techniques Conference Conference Papers},
  pages={1--12},
  year={2025}
}

@article{wang2025ati,
  title={Ati: Any trajectory instruction for controllable video generation},
  author={Wang, Angtian and Huang, Haibin and Fang, Jacob Zhiyuan and Yang, Yiding and Ma, Chongyang},
  journal={arXiv preprint arXiv:2505.22944},
  year={2025}
}

@article{geyer2023tokenflow,
  title={Tokenflow: Consistent diffusion features for consistent video editing},
  author={Geyer, Michal and Bar-Tal, Omer and Bagon, Shai and Dekel, Tali},
  journal={arXiv preprint arXiv:2307.10373},
  year={2023}
}

@inproceedings{qi2023fatezero,
  title={Fatezero: Fusing attentions for zero-shot text-based video editing},
  author={Qi, Chenyang and Cun, Xiaodong and Zhang, Yong and Lei, Chenyang and Wang, Xintao and Shan, Ying and Chen, Qifeng},
  booktitle={Proceedings of the IEEE/CVF International Conference on Computer Vision},
  pages={15932--15942},
  year={2023}
}

@inproceedings{yang2023rerender,
  title={Rerender a video: Zero-shot text-guided video-to-video translation},
  author={Yang, Shuai and Zhou, Yifan and Liu, Ziwei and Loy, Chen Change},
  booktitle={SIGGRAPH Asia 2023 Conference Papers},
  pages={1--11},
  year={2023}
}

@inproceedings{wang2025videodirector,
  title={Videodirector: Precise video editing via text-to-video models},
  author={Wang, Yukun and Wang, Longguang and Ma, Zhiyuan and Hu, Qibin and Xu, Kai and Guo, Yulan},
  booktitle={Proceedings of the IEEE/CVF Conference on Computer Vision and Pattern Recognition},
  pages={2589--2598},
  year={2025}
}

@inproceedings{li2024vidtome,
  title={Vidtome: Video token merging for zero-shot video editing},
  author={Li, Xirui and Ma, Chao and Yang, Xiaokang and Yang, Ming-Hsuan},
  booktitle={Proceedings of the IEEE/CVF Conference on Computer Vision and Pattern Recognition},
  pages={7486--7495},
  year={2024}
}

@inproceedings{zhou2023propainter,
  title={Propainter: Improving propagation and transformer for video inpainting},
  author={Zhou, Shangchen and Li, Chongyi and Chan, Kelvin CK and Loy, Chen Change},
  booktitle={Proceedings of the IEEE/CVF international conference on computer vision},
  pages={10477--10486},
  year={2023}
}

@article{mildenhall2022nerf,
  title={Nerf: Representing scenes as neural radiance fields for view synthesis},
  author={Mildenhall, Ben and Srinivasan, Pratul P and Tancik, Matthew and Barron, Jonathan T and Ramamoorthi, Ravi and Ng, Ren},
  journal={Communications of the ACM},
  volume={65},
  number={1},
  pages={99--106},
  year={2022},
  publisher={ACM New York, NY, USA}
}

@article{kerbl20233d,
  title={3d gaussian splatting for real-time radiance field rendering.},
  author={Kerbl, Bernhard and Kopanas, Georgios and Leimk{\"u}hler, Thomas and Drettakis, George},
  journal={ACM Trans. Graph.},
  volume={42},
  number={4},
  pages={139:1--139:14},
  year={2023}
}

@inproceedings{wu20244d,
  title={4d gaussian splatting for real-time dynamic scene rendering},
  author={Wu, Guanjun and Yi, Taoran and Fang, Jiemin and Xie, Lingxi and Zhang, Xiaopeng and Wei, Wei and Liu, Wenyu and Tian, Qi and Wang, Xinggang},
  booktitle={Proceedings of the IEEE/CVF conference on computer vision and pattern recognition},
  pages={20310--20320},
  year={2024}
}

@inproceedings{wu2024reconfusion,
  title={Reconfusion: 3d reconstruction with diffusion priors},
  author={Wu, Rundi and Mildenhall, Ben and Henzler, Philipp and Park, Keunhong and Gao, Ruiqi and Watson, Daniel and Srinivasan, Pratul P and Verbin, Dor and Barron, Jonathan T and Poole, Ben and others},
  booktitle={Proceedings of the IEEE/CVF conference on computer vision and pattern recognition},
  pages={21551--21561},
  year={2024}
}

@inproceedings{lin2026vista4d,
  title={Vista4D: Video Reshooting with 4D Point Clouds},
  author={Lin, Kuan Heng and Liu, Zhizheng and Salamanca, Pablo and Kant, Yash and Burgert, Ryan and Xu, Yuancheng and Namekata, Koichi and Zhao, Yiwei and Zhou, Bolei and Goldblum, Micah and others},
  booktitle={Proceedings of the IEEE/CVF Conference on Computer Vision and Pattern Recognition},
  pages={32671--32682},
  year={2026}
}

@article{yang2026neoverse,
  title={NeoVerse: Enhancing 4D World Model with in-the-wild Monocular Videos},
  author={Yang, Yuxue and Fan, Lue and Shi, Ziqi and Peng, Junran and Wang, Feng and Zhang, Zhaoxiang},
  journal={arXiv preprint arXiv:2601.00393},
  year={2026}
}

@article{wu2024recent,
  title={Recent advances in 3d gaussian splatting},
  author={Wu, Tong and Yuan, Yu-Jie and Zhang, Ling-Xiao and Yang, Jie and Cao, Yan-Pei and Yan, Ling-Qi and Gao, Lin},
  journal={Computational Visual Media},
  volume={10},
  number={4},
  pages={613--642},
  year={2024},
  publisher={TUP}
}

@article{wan2025wan,
  title={Wan: Open and advanced large-scale video generative models},
  author={Wan, Team and Wang, Ang and Ai, Baole and Wen, Bin and Mao, Chaojie and Xie, Chen-Wei and Chen, Di and Yu, Feiwu and Zhao, Haiming and Yang, Jianxiao and others},
  journal={arXiv preprint arXiv:2503.20314},
  year={2025}
}

@inproceedings{yao2025uni4d,
  title={Uni4d: Unifying visual foundation models for 4d modeling from a single video},
  author={Yao, David Yifan and Zhai, Albert J and Wang, Shenlong},
  booktitle={Proceedings of the Computer Vision and Pattern Recognition Conference},
  pages={1116--1126},
  year={2025}
}

@article{lu2025trackingworld,
  title={TrackingWorld: World-centric Monocular 3D Tracking of Almost All Pixels},
  author={Lu, Jiahao and Xiong, Weitao and Deng, Jiacheng and Li, Peng and Huang, Tianyu and Dou, Zhiyang and Lin, Cheng and Yeung, Sai-Kit and Liu, Yuan},
  journal={arXiv preprint arXiv:2512.08358},
  year={2025}
}

@article{liu2025trace,
  title={Trace anything: Representing any video in 4d via trajectory fields},
  author={Liu, Xinhang and Xiao, Yuxi and Chen, Donny Y and Feng, Jiashi and Tai, Yu-Wing and Tang, Chi-Keung and Kang, Bingyi},
  journal={arXiv preprint arXiv:2510.13802},
  year={2025}
}

@inproceedings{karaev2024cotracker,
  title={Cotracker: It is better to track together},
  author={Karaev, Nikita and Rocco, Ignacio and Graham, Benjamin and Neverova, Natalia and Vedaldi, Andrea and Rupprecht, Christian},
  booktitle={European conference on computer vision},
  pages={18--35},
  year={2024},
  organization={Springer}
}

@inproceedings{xiao2024spatialtracker,
  title={Spatialtracker: Tracking any 2d pixels in 3d space},
  author={Xiao, Yuxi and Wang, Qianqian and Zhang, Shangzhan and Xue, Nan and Peng, Sida and Shen, Yujun and Zhou, Xiaowei},
  booktitle={Proceedings of the IEEE/CVF Conference on Computer Vision and Pattern Recognition},
  pages={20406--20417},
  year={2024}
}

@article{yang2024cogvideox,
  title={Cogvideox: Text-to-video diffusion models with an expert transformer},
  author={Yang, Zhuoyi and Teng, Jiayan and Zheng, Wendi and Ding, Ming and Huang, Shiyu and Xu, Jiazheng and Yang, Yuanming and Hong, Wenyi and Zhang, Xiaohan and Feng, Guanyu and others},
  journal={arXiv preprint arXiv:2408.06072},
  year={2024}
}

@inproceedings{chen2025perception,
  title={Perception-as-control: Fine-grained controllable image animation with 3d-aware motion representation},
  author={Chen, Yingjie and Men, Yifang and Yao, Yuan and Cui, Miaomiao and Bo, Liefeng},
  booktitle={Proceedings of the IEEE/CVF International Conference on Computer Vision},
  pages={14380--14389},
  year={2025}
}

@article{huang2025voyager,
  title={Voyager: Long-range and world-consistent video diffusion for explorable 3d scene generation},
  author={Huang, Tianyu and Zheng, Wangguandong and Wang, Tengfei and Liu, Yuhao and Wang, Zhenwei and Wu, Junta and Jiang, Jie and Li, Hui and Lau, Rynson and Zuo, Wangmeng and others},
  journal={ACM Transactions on Graphics (TOG)},
  volume={44},
  number={6},
  pages={1--15},
  year={2025},
  publisher={ACM New York, NY, USA}
}

@article{li2025flashworld,
  title={FlashWorld: High-quality 3D Scene Generation within Seconds},
  author={Li, Xinyang and Wang, Tengfei and Gu, Zixiao and Zhang, Shengchuan and Guo, Chunchao and Cao, Liujuan},
  journal={arXiv preprint arXiv:2510.13678},
  year={2025}
}
\bibliographystyle{plainnat}

\clearpage

\appendix

\addtocontents{toc}{\protect\setcounter{tocdepth}{3}}
\begingroup
\renewcommand{\contentsname}{Appendix Contents}
\setcounter{tocdepth}{3}
\makeatletter
\renewcommand*\l@section[2]{%
  \ifnum \c@tocdepth >\z@
    \addpenalty\@secpenalty
    \addvspace{0.4em \@plus\p@}%
    \setlength\@tempdima{1.5em}%
    \begingroup
      \parindent \z@ \rightskip \@pnumwidth
      \parfillskip -\@pnumwidth
      \leavevmode \bfseries
      \advance\leftskip\@tempdima
      \hskip -\leftskip
      #1\nobreak\hfil
      \nobreak\hb@xt@\@pnumwidth{\hss #2\kern-\p@\kern\p@}\par
    \endgroup
  \fi}
\makeatother
\tableofcontents
\endgroup
\clearpage

\section{Additional Method Details}
\label{app:prex_details}

This appendix specifies the upstream representation, projection rules, and proxy-task construction summarized in Section~\ref{sec:method_overview}. PREX operates on an editable 4D proxy whose points can be associated with source observations.

\subsection{Video-to-4D Reconstruction and Proxy Representation}
\label{app:reconstruction}

\paragraph{Modular reconstruction.}
Given $V=\{I_t\}_{t=1}^{T}$, we estimate object-level semantic priors $\mathcal O$, per-frame depth $\mathcal D=\{D_t\}$, temporally consistent instance masks $\mathcal M=\{M_t^{ins}\}$, dense point tracks $\mathcal Q$, and camera parameters $\{K_t,E_t\}$. A 4D fusion module produces
\begin{equation}
\mathcal S=F_{\mathrm{4D}}\!\left(
V,\mathcal O,\mathcal D,\mathcal M,\mathcal Q,
\{K_t,E_t\}_{t=1}^{T}
\right).
\label{eq:4d_fusion}
\end{equation}
Semantic discovery, point tracking, monocular depth, open-vocabulary instance segmentation, and temporal mask propagation are separate modules. Instance identities are propagated before fusion to maintain consistent instance identities across frames. We additionally store aligned tracks $\mathcal T$ with 2D projections, 3D positions, visibility, and instance identity. The downstream PREX controls depend only on the standardized proxy and can therefore use improved reconstruction modules without changing the adapter architecture.

\paragraph{Static--dynamic factorization.}
We represent the proxy as
\begin{equation}
\mathcal S=\left\{
P^{sta},C^{sta},
\{P_t^{dyn},C_t^{dyn},A_t^{dyn}\}_{t=1}^{T},
\{K_t,E_t,I_t\}_{t=1}^{T}
\right\},
\label{eq:4d_proxy}
\end{equation}
where static points are stored once, while dynamic positions, colors, and instance IDs are time dependent. This factorization permits camera edits and object-level translation, rotation, scaling, trajectory editing, removal, or dynamicity changes. We write the edited proxy as
\begin{equation}
\mathcal S'=\mathcal E(\mathcal S;\Theta_{cam},\Theta_{obj}),
\label{eq:edited_scene}
\end{equation}
and keep source and edited geometry in a shared world coordinate system.

\paragraph{Projection interface.}
For target camera $\Pi_t$, z-buffer projection first produces intermediate buffers
\begin{equation}
R_t^{proj}=\{\bar C_t^{rgb},D_t,I_t^{ins},H_t,
g_t^{cov},g_t^{pur},g_t^{std}\},
\label{eq:projection_buffers}
\end{equation}
including a coarse rendered color, depth, instance identity, an unsupported-projection indicator, and statistics used for confidence. These are not all injected into the diffusion backbone. Instead, the procedures below convert them and the source video into the compact final control $\mathcal R_t$ in Eq.~\ref{eq:framewise_control_state}.

\subsection{Observation Verification and Appearance Rendering}
\label{app:observation_details}

For target pixel $x$, let $X'_t(x)$ be the visible edited 4D point. Static points remain in the shared world frame. For a dynamic instance, we apply the inverse user edit and the source-time object transform to obtain the corresponding source-space point $X_{t\rightarrow r}(x)$. Its source projection is
\begin{equation}
u_{t,r}(x)=\pi_r\!\left(X_{t\rightarrow r}(x)\right).
\label{eq:source_projection}
\end{equation}
For candidate frame $r\in\mathcal W_t$, validity is the conjunction of image-domain visibility, depth agreement, and instance agreement:
\begin{equation}
\begin{aligned}
\nu_{t,r}(x)={}&
\ind{u_{t,r}(x)\in\Omega_r}
\ind{\left|z_r(X_{t\rightarrow r}(x))-D_r(u_{t,r}(x))\right|<\epsilon_d}\\
&\cdot
\ind{a_r(u_{t,r}(x))=a(X'_t(x))}.
\end{aligned}
\label{eq:candidate_validity}
\end{equation}
Here $z_r$ is the projected point depth, $D_r$ is the source depth map, and $a_r$ is the temporally propagated instance label. The first term rejects off-image points, the second rejects samples that fail the estimated depth-consistency test, and the third rejects samples with mismatched instance labels.

Equation~\ref{eq:preserve_mask} marks a target pixel as Preserve when at least one candidate is valid. We then rank valid candidates with Eq.~\ref{eq:view_time_score}. The best observation supplies $C_t^{rgb}(x)$; when several nearby samples have similar scores and stable support, local weighted compositing reduces point-projection holes. When no candidate is valid, PREX either retains a weak coarse prior $\bar C_t^{rgb}(x)$ or leaves the control unresolved, always with zero confidence.

\subsection{Scene Extent, Region Roles, and Confidence}
\label{app:role_details}

The Preserve mask alone does not distinguish an in-scene disocclusion from content outside the captured field of view. We compute the original scene-extent mask as
\begin{equation}
M_t^S=\mathrm{Close}\!\left(M_t^{geo}\lor M_t^{fov}\right),
\label{eq:scene_extent_mask}
\end{equation}
where $M_t^{geo}$ is the original reconstructed proxy rendered from $\Pi_t$ and $M_t^{fov}$ is the union of source-camera frusta projected into the target view. Morphological closing reduces spurious expansion labels caused by small reconstruction cracks. Combining $M_t^S$ with observation support yields the Reveal and Expand masks in Eq.~\ref{eq:reveal_expand_masks}. By construction,
\begin{equation}
M_t^P\cup M_t^R\cup M_t^E=\mathbf 1,
\qquad
M_t^P\cap M_t^R=M_t^P\cap M_t^E=M_t^R\cap M_t^E=\emptyset.
\label{eq:role_partition}
\end{equation}

The low-level hole mask $H_t$ only records the absence of directly projected proxy support. It is not a semantic editing role: a hole within $M_t^S$ is Reveal, whereas a hole outside $M_t^S$ is Expand. Non-hole pixels can also receive low confidence when their projected support is unstable.

Before view-dependent attenuation, geometric confidence is
\begin{equation}
c_t^{geo}=g_t^{cov}g_t^{pur}\exp(-g_t^{std}/\tau),
\label{eq:geometric_confidence}
\end{equation}
where coverage measures projected sample support, purity measures local instance consistency, and depth variation penalizes unstable surfaces and boundaries. Equation~\ref{eq:calibrated_confidence} additionally attenuates this value when source and target viewing directions disagree. Unsupported Reveal and Expand pixels are assigned zero confidence. The masks therefore communicate the required behavior, while confidence provides a continuous reliability signal for the adapter.

\subsection{Detailed Proxy-task Construction}
\label{app:proxy_tasks}

Each proxy task changes only the evidence available in the conditioning input while keeping the raw video $V$ as the reconstruction target. Our 10,000 training pairs contain 5,000 Preserve-oriented reconstruction pairs, 2,500 Reveal-proxy pairs, and 2,500 Expand-proxy pairs. The same source clip may appear under different condition types, but each sampled pair has one independently constructed control sequence. All variants share the same adapter, frozen video backbone, and flow-matching objective.

\paragraph{Condition and supervision.}
For every pair, we reconstruct the unedited clip into a full proxy $\mathcal S$ and apply a support-degradation operator to obtain $\tilde{\mathcal S}$. The RGB cue and confidence map are constructed from $\tilde{\mathcal S}$, and the role masks are constructed as described below. PREX receives these controls together with the original first frame required by the Wan2.1-I2V backbone. The target is the VAE latent of the unchanged RGB clip, and the standard flow-matching velocity loss is applied over all latent locations.

\paragraph{Preserve-oriented reconstruction.}
We retain the original proxy and use the original or a nearby target view,
\begin{equation}
\tilde{\mathcal S}=\mathcal S,\qquad
\tilde\Pi_t\approx\Pi_t,\qquad M_t^P\approx\mathbf 1.
\label{eq:preserve_task}
\end{equation}
This variant aligns source-backed appearance latents with the frozen video model and teaches high-fidelity preservation.

\paragraph{Reveal-oriented disocclusion.}
We erase static support that is occluded by moving foreground geometry. We sample an active-window ratio $s\sim\mathcal U(0.30,0.60)$ and set $T_{act}=\lceil sT\rceil$. During the first $T_{act}$ frames, a static point is marked for removal when its projected depth lies behind a dynamic point by more than $0.05$ in normalized depth. Dynamic support is dilated with a two-pixel splat radius before this comparison to reduce leakage of withheld background through flying edges and narrow foreground gaps. To avoid framewise flicker, the erased set is accumulated during the active window and then frozen:
\begin{equation}
M_{erase}^{t}=M_{erase}^{t-1}\cup M_{occ}^{t},\quad t\leq T_{act},
\qquad
\tilde{\mathcal S}=\mathcal S\setminus P_{erase}^{sta}.
\label{eq:erase_accumulation}
\end{equation}
The removed support lies inside the captured scene extent, so its unsupported projection is labeled Reveal. Reconstructing the original video teaches the adapter to suppress invalid copied foreground appearance and fill disocclusions from surrounding spatial and temporal context.

\paragraph{Expand-oriented partial observation.}
We sample a keep ratio $\rho\sim\mathcal U(0.15,0.30)$, set $T_p=\lceil\rho T\rceil$, and retain only static points observed in the source prefix:
\begin{equation}
P_{keep}^{sta}=\{X\in P^{sta}\mid
\exists t\leq T_p:\mathrm{Visible}(X,\Pi_t)=1\}.
\label{eq:prefix_static_keep}
\end{equation}
Dense dynamic points are retained so that the perturbation primarily changes the observed scene extent rather than removing the foreground subject. Rendering this partial proxy at later source cameras creates unsupported regions outside the retained observation footprint; these pixels are labeled Expand. The task therefore teaches extrapolation from a restricted field of observation while preserving the source clip as supervision.

\paragraph{Role-mask construction.}
The following implementation details supplement the scene-extent construction in Appendix~\ref{app:role_details} and the role definitions in Eq.~\ref{eq:reveal_expand_masks}. The full and support-degraded proxies are rendered with the same z-buffered rasterizer. Their support difference supplies the intended Reveal or Expand region for the sampled task. Tracking gaps and unsupported components created behind dynamic instances are assigned to Reveal. For remaining unsupported components, we refine the scene-extent assignment using 8-connectivity: a component touching an eight-pixel image-border band is treated as outside the observed extent and assigned to Expand, while an interior component is treated as inside the extent and assigned to Reveal. Preserve contains pixels with valid source-observation support after the visibility, depth, and instance checks in Appendix~\ref{app:observation_details}. Pixels that fail these checks are removed from Preserve and classified by the same component rule. The three masks are consequently disjoint and cover the complete target frame.

\paragraph{Joint sampling and robustness.}
During the curriculum phase, variants are mixed within one optimization process,
\begin{equation}
q\sim\{P,R,E\},\qquad
\tilde{\mathcal E}=\tilde{\mathcal E}_q,
\label{eq:proxy_variant_sampling}
\end{equation}
to encourage compatible role-specific behavior at region boundaries. Training and inference use the same rasterizer, observation-validity checks, confidence computation, and Preserve/Reveal/Expand partition, including the component-based scene-extent refinement above. Training obtains the three regimes through controlled support degradation, whereas inference obtains them from actual camera/object edits and source-observation validity. We also perturb dynamic geometry to mimic flying edges and instance-shape errors, and jitter appearance retrieval to discourage unconditional copying. Perturbations are sampled per scene with deterministic seeds for reproducibility.

\section{Evaluation Metrics}
\label{app:metrics}

\subsection{PREBench Metrics}
\label{app:prebench_metrics}

PREBench evaluates faithful 4D video editing with region-aware metrics over three edit-induced regions: \textit{Preserve}, \textit{Reveal}, and \textit{Expand}. 
Given a generated video $\hat{I}=\{\hat{I}_t\}_{t=1}^{T}$, the source-backed preserve reference $I^{p}=\{I^{p}_t\}_{t=1}^{T}$, the coarse rendered evidence or Reveal ghost reference $I^{g}=\{I^{g}_t\}_{t=1}^{T}$, and the exported RGB hint $I^{h}=\{I^{h}_t\}_{t=1}^{T}$, PREBench uses the edit-region masks $M^P_t$, $M^R_t$, $M^E_t$, and $M^{D}_t$ for preserve, reveal, expand, and dynamic-preserve regions, respectively. An exported mask $U^h_t$ identifies hint pixels that are unsupported or otherwise invalid for direct copying.

All image values are normalized to $[0,1]$ unless otherwise specified. 
Metrics are first computed per case and then averaged over all valid cases. 
If the corresponding region mask is empty for a case, the metric is ignored for that case during aggregation. 
Lower values are better for all nine component metrics. The aggregate PREBench Score defined below is higher-is-better.

\subsubsection{Preserve Metrics}

Preserve regions are source-backed pixels that should remain faithful to the input observation. 
PREBench evaluates both perceptual appearance drift and temporal drift within these regions.

\paragraph{P-LPIPS.}
P-LPIPS measures perceptual drift in preserve regions using LPIPS. 
Since LPIPS is an image-level perceptual metric, we apply the preserve mask by compositing both images onto a neutral background:
\begin{equation}
\tilde{I}_t^{M} = M_t \odot I_t + (1-M_t)\odot c,
\end{equation}
where $c$ is a constant neutral RGB color. 
The preserve LPIPS score is
\begin{equation}
\mathrm{P\text{-}LPIPS}
=
\frac{1}{|\mathcal{T}_P|}
\sum_{t \in \mathcal{T}_P}
\mathrm{LPIPS}
\left(
\tilde{\hat{I}}_t^{M^P_t},
\tilde{I}^{p,M^P_t}_t
\right),
\end{equation}
where $\mathcal{T}_P=\{t: |M^P_t|>0\}$.

\paragraph{P-DISTS.}
P-DISTS similarly measures perceptual and structural distortion in preserve regions using DISTS:
\begin{equation}
\mathrm{P\text{-}DISTS}
=
\frac{1}{|\mathcal{T}_P|}
\sum_{t \in \mathcal{T}_P}
\mathrm{DISTS}
\left(
\tilde{\hat{I}}_t^{M^P_t},
\tilde{I}^{p,M^P_t}_t
\right).
\end{equation}
This metric complements LPIPS by emphasizing structural consistency in source-backed regions.

\paragraph{P-TempDrift.}
P-TempDrift measures whether the temporal change in preserve regions matches the temporal change in the source-backed reference. 
It is computed from the difference between adjacent-frame residuals:
\begin{equation}
\mathrm{P\text{-}TempDrift}
=
\frac{1}{|\mathcal{T}_{P}^{\Delta}|}
\sum_{t=2}^{T}
\frac{1}{|M^P_t \cap M^P_{t-1}|}
\sum_{x \in M^P_t \cap M^P_{t-1}}
\left|
(\hat{I}_t(x)-\hat{I}_{t-1}(x))
-
(I^p_t(x)-I^p_{t-1}(x))
\right|,
\end{equation}
where only frame pairs with non-empty overlapping preserve masks are included. 
This metric penalizes temporal drift even when individual frames remain visually plausible.

\paragraph{P-Dyn-LPIPS.}
P-Dyn-LPIPS measures perceptual preservation specifically on dynamic source-backed regions, such as edited or tracked moving objects that should preserve their original appearance:
\begin{equation}
\mathrm{P\text{-}Dyn\text{-}LPIPS}
=
\frac{1}{|\mathcal{T}_D|}
\sum_{t \in \mathcal{T}_D}
\mathrm{LPIPS}
\left(
\tilde{\hat{I}}_t^{M^D_t},
\tilde{I}^{p,M^D_t}_t
\right),
\end{equation}
where $\mathcal{T}_D=\{t: |M^D_t|>0\}$. 
This isolates preservation quality on dynamic preserve masks instead of averaging it with static background regions.

\subsubsection{Reveal Metrics}

Reveal regions correspond to newly disoccluded pixels that are inside the original scene extent but are not directly supported by valid source observations. 
These regions should be completed coherently while avoiding leakage from invalid coarse evidence.

\paragraph{R-Ghost.}
R-Ghost measures how strongly the generated reveal region resembles the ghost reference $I^g$, i.e., the coarse or invalid rendered evidence. 
We first compute the masked mean absolute error:
\begin{equation}
\mathrm{MAE}_R
=
\frac{1}{|\Omega_R|}
\sum_{t,x \in M^R_t}
\left|
\hat{I}_t(x)-I^g_t(x)
\right|,
\end{equation}
and convert it into a similarity score:
\begin{equation}
\mathrm{R\text{-}Ghost}
=
\exp\left(
-\frac{\mathrm{MAE}_R}{\sigma}
\right).
\end{equation}
A high value indicates strong similarity to the ghost reference, providing a signal of potential ghost leakage. 
We use $\sigma=0.18$ in our implementation.

\paragraph{R-Seam.}
R-Seam measures color discontinuity along the boundary between reveal and preserve regions. 
For each frame, we define an inner reveal boundary band and an outer preserve boundary band:
\begin{equation}
B^R_t = M^R_t \cap \operatorname{Dilate}(M^P_t, r),
\end{equation}
\begin{equation}
B^P_t = M^P_t \cap \operatorname{Dilate}(M^R_t, r),
\end{equation}
where $r$ is the boundary radius. 
The seam score is the mean RGB difference between the two boundary bands:
\begin{equation}
\mathrm{R\text{-}Seam}
=
\frac{1}{|\mathcal{T}_{R,S}|}
\sum_{t \in \mathcal{T}_{R,S}}
\left\|
\mu(\hat{I}_t, B^R_t)
-
\mu(\hat{I}_t, B^P_t)
\right\|_1 ,
\end{equation}
where $\mu(I,M)$ denotes the mean RGB color of image $I$ over mask $M$. 
This metric provides a color-based diagnostic of transitions between revealed content and preserved observations.

\subsubsection{Expand Metrics}

Expand regions correspond to newly visible out-of-view content caused by camera reshooting or view expansion. 
These pixels have no direct source support and should be synthesized temporally coherently without simply copying or stretching old boundary evidence.

\paragraph{E-Temp.}
E-Temp measures temporal inconsistency after compensating for camera and scene motion. 
For each adjacent pair, a fixed optical-flow estimator computes backward flow $F_{t\rightarrow t-1}$ from $\hat I_t$ to $\hat I_{t-1}$. 
We warp the preceding frame and its Expand mask into frame $t$ and retain only forward--backward-consistent, non-occluded correspondences:
\begin{equation}
V_t^E
=
M_t^E
\cap
\mathbf{1}\!\left[\mathcal W(M_{t-1}^E,F_{t\rightarrow t-1})>0.5\right]
\cap O_t,
\label{eq:e_temp_valid}
\end{equation}
where $\mathcal W$ denotes bilinear warping and $O_t$ is the validity mask obtained from forward--backward flow consistency. 
Specifically, $O_t(x)=1$ when the forward--backward flow discrepancy is smaller than $0.01$ times the summed flow magnitude plus $0.5$ pixels. 
Let $\Phi(\cdot)$ denote the normalized final-layer patch tokens of a fixed DINOv2 ViT-B/14 encoder, with the flow and validity mask resized to the feature resolution. 
The photometric and feature residuals are
\begin{align}
e_t^{\mathrm{rgb}}
&=
\frac{1}{|V_t^E|}
\sum_{x\in V_t^E}
\rho\!\left(
\hat I_t(x)-\mathcal W(\hat I_{t-1},F_{t\rightarrow t-1})(x)
\right),
\\
e_t^{\mathrm{feat}}
&=
\frac{1}{|V_{t,\Phi}^E|}
\sum_{u\in V_{t,\Phi}^E}
\left[
1-cos\!\left(
\Phi(\hat I_t)(u),
\mathcal W(\Phi(\hat I_{t-1}),F_{t\rightarrow t-1})(u)
\right)
\right],
\end{align}
where $\rho(z)=\operatorname{mean}_c\sqrt{z_c^2+\epsilon^2}$ is the Charbonnier penalty. 
The final score is
\begin{equation}
\mathrm{E\text{-}Temp}
=
\frac{1}{|\mathcal T_E^{\Delta}|}
\sum_{t\in\mathcal T_E^{\Delta}}
\left(
\lambda_T e_t^{\mathrm{rgb}}
+(1-\lambda_T)e_t^{\mathrm{feat}}
\right),
\label{eq:e_temp}
\end{equation}
where $\mathcal T_E^{\Delta}=\{t\in\{2,\ldots,T\}:|V_t^E|>0,|V_{t,\Phi}^E|>0\}$. 
We use $\epsilon=10^{-3}$ and $\lambda_T=0.5$ for all methods. 
Lower E-Temp therefore indicates less unexplained flicker or feature drift after motion compensation; newly appearing or occluded pixels are not scored.

\paragraph{E-Seam.}
E-Seam measures excess feature discontinuity across direct Expand--Preserve interfaces relative to local variation on either side. 
Unlike a difference between band-wise mean colors, the metric pairs samples across the interface and discounts the local variation already present on each side. 
For frame $t$, let
\begin{equation}
\Gamma_t^{E,P}
=
\partial M_t^E
\cap
\operatorname{Dilate}(M_t^P,1)
\end{equation}
be the valid direct interface. 
For each $p\in\Gamma_t^{E,P}$, the signed-distance transform provides a unit normal $n_p$ pointing from Preserve to Expand. 
We sample $z_p^E=p+\delta n_p$ and $z_p^P=p-\delta n_p$, together with the corresponding interior samples $z_{p,+}^E=p+2\delta n_p$ and $z_{p,-}^P=p-2\delta n_p$. 
Here $\delta=\max(1,\operatorname{round}(0.005\min(H,W)))$, and points whose four samples are not valid are discarded.

Let $\psi(I,z)$ be a fixed multi-scale descriptor formed by normalized CIELAB statistics over $7$- and $15$-pixel neighborhoods and the aligned DINOv2 patch feature, and let
\begin{equation}
d_t(a,b)
=
\lambda_S
\left\|\psi_{\mathrm{Lab}}(\hat I_t,a)-\psi_{\mathrm{Lab}}(\hat I_t,b)\right\|_1
+
(1-\lambda_S)
\left[1-\cos\!\left(\psi_{\mathrm{DINO}}(\hat I_t,a),\psi_{\mathrm{DINO}}(\hat I_t,b)\right)\right].
\end{equation}
The excess discontinuity at $p$ is
\begin{equation}
s_t(p)
=
\left[
d_t(z_p^E,z_p^P)
-
\frac{1}{2}
\left(
d_t(z_p^E,z_{p,+}^E)
+
d_t(z_p^P,z_{p,-}^P)
\right)
\right]_+ .
\end{equation}
We aggregate robustly over the interface:
\begin{equation}
\mathrm{E\text{-}Seam}
=
\frac{1}{|\mathcal T_{E,S}|}
\sum_{t\in\mathcal T_{E,S}}
\operatorname{TrimMean}_{10\%}
\{s_t(p):p\in\Gamma_t^{E,P}\},
\label{eq:e_seam}
\end{equation}
where $\mathcal T_{E,S}$ contains frames with at least one valid direct interface sample, and $\lambda_S=0.5$. 
Frames in which Reveal or another unsupported region separates Expand from Preserve are excluded. 
A lower value indicates less cross-interface discontinuity relative to the local variation of the generated content.

\paragraph{E-Copy.}
E-Copy diagnoses signatures of unreliable evidence reuse through two complementary signals: similarity to an invalid RGB hint and patch-level similarity to the source-backed field-of-view boundary. 

For an image pair $A,B$, let $q(A,B;x,y)\in[0,1]$ denote the maximum zero-mean normalized cross-correlation between multi-scale patches centered at $x$ and $y$, mapped from $[-1,1]$ to $[0,1]$. 
We use patch widths $\{7,15,31\}$ at the evaluation resolution and source-patch scale factors $\{0.75,1,1.25\}$ so that repeated and moderately stretched textures can both be detected.

The invalid-hint reuse score is
\begin{equation}
S_t^{\mathrm{hint}}
=
\frac{1}{|Q_t^h|}
\sum_{x\in Q_t^h}
q(\hat I_t,I_t^h;x,x),
\qquad
Q_t^h=M_t^E\cap U_t^h,
\label{eq:e_hint_trust}
\end{equation}
and is defined only when $|Q_t^h|>0$. 
This term is high when the output reproduces the spatially aligned but invalid RGB hint.

For boundary reuse, define the source-backed boundary strip
\begin{equation}
C_t^P
=
M_t^P\cap\operatorname{Dilate}(M_t^E,r_b),
\qquad
r_b=\operatorname{round}(0.02\min(H,W)).
\end{equation}
For an Expand patch center $x$, $b_t(x)$ is its closest point on the known source field-of-view boundary and
$\mathcal N_t(x)=\{y\in C_t^P:\|y-b_t(x)\|_2\le r_b\}$ is the corresponding candidate set. 
To avoid labeling homogeneous regions as copies, we weight each source patch by its normalized texture energy $a_t(y)=\min(1,\operatorname{Var}(\nabla P_{15}(I_t^p,y))/\nu)$. 
The normalization $\nu$ is fixed to the 75th percentile of gradient variance over source-boundary patches in the benchmark and is computed without using generated outputs. 
The boundary-copy score is
\begin{equation}
S_t^{\mathrm{bdry}}
=
\frac{1}{|Q_t^E|}
\sum_{x\in Q_t^E}
\max_{y\in\mathcal N_t(x)}
a_t(y)q(\hat I_t,I_t^p;x,y),
\label{eq:e_boundary_copy}
\end{equation}
where $Q_t^E$ contains Expand patch centers with valid candidates and complete patches. 
Let $\mathcal K_t\subseteq\{\mathrm{hint},\mathrm{bdry}\}$ contain the components available in frame $t$. 
The components are combined by a fixed soft-OR,
\begin{equation}
S_t^{\mathrm{copy}}
=
1-
\prod_{k\in\mathcal K_t}
\left(1-S_t^k\right),
\end{equation}
so an unavailable component is omitted rather than treated as an observed failure. 
The final score is
\begin{equation}
\mathrm{E\text{-}Copy}
=
\frac{1}{|\mathcal T_{E,C}|}
\sum_{t\in\mathcal T_{E,C}}
S_t^{\mathrm{copy}},
\label{eq:e_copy}
\end{equation}
where $\mathcal T_{E,C}$ contains frames with at least one valid component. 
Cases without an invalid hint and without an informative source-boundary strip are marked unavailable for E-Copy. 
Higher E-Copy indicates stronger similarity signatures associated with evidence reuse; the two component scores are retained for diagnostic analysis. 
E-Copy targets signatures of copying, stretching, and invalid-hint reuse.

\subsubsection{Aggregate PREBench Score}
The nine component metrics have different numerical scales and therefore cannot be averaged directly. We first convert each error metric $x_m$ to a fixed 0--100 quality score:
\begin{equation}
q_m(x_m)=100\,2^{-x_m/h_m},
\end{equation}
where $h_m$ is the error value mapped to 50 points. In the order P-LPIPS, P-DISTS, P-TempDrift, P-Dyn-LPIPS, R-Ghost, R-Seam, E-Temp, E-Seam, and E-Copy, the calibrated anchors used to compute the reported scores are
\begin{equation}
(h_m)=(0.25,\,0.15,\,0.08,\,0.25,\,0.20,\,0.08,\,0.03,\,0.08,\,0.30).
\end{equation}
Thus, zero error maps to 100, $h_m$ maps to 50, and $2h_m$ maps to 25. These reference scales were calibrated from human severity ratings of moderate target degradation, equally sampled from the PREBench test split and an independent calibration set, using Eq.~\ref{eq:human_calibrated_anchor}, as reported in Section~\ref{app:metric_validation} and Table~\ref{tab:anchor_calibration_appendix}. The same calibrated anchors were applied to every method when computing the aggregate scores reported throughout the paper.

We then average normalized qualities within each evidence role:
\begin{equation}
S_P=\frac{1}{4}\sum_{m\in\mathcal M_P}q_m,
\qquad
S_R=\frac{1}{2}\sum_{m\in\mathcal M_R}q_m,
\qquad
S_E=\frac{1}{3}\sum_{m\in\mathcal M_E}q_m,
\end{equation}
where $\mathcal M_P$, $\mathcal M_R$, and $\mathcal M_E$ contain the four Preserve, two Reveal, and three Expand metrics, respectively. The final score is
\begin{equation}
\mathrm{PREBench\ Score}=\frac{S_P+S_R+S_E}{3}.
\end{equation}
This region-balanced aggregation prevents Preserve from receiving extra weight merely because it has more component metrics. If a case contains no dynamic-preserve pixels, P-Dyn-LPIPS is marked not applicable and omitted from $S_P$ for that case; a required but unavailable metric does not receive an imputed score.

\subsection{VBench Metrics}
\label{app:vbench_metrics}

We assess image-to-video generation quality with the VBench Image-to-Video evaluation suite, referred to as VBench. 
For each generated clip, we follow the official VBench evaluation protocol: the input conditioning image and the generated video are jointly fed into the evaluation pipeline, which returns a set of learned and human-aligned scores for measuring both perceptual video quality and image-video consistency.

In our experiments, we report six VBench dimensions. 
The first group measures general video quality, including frame-level fidelity, aesthetics, motion strength, temporal smoothness, and temporal consistency of background and subject. 
The second group evaluates image-to-video consistency, measuring whether the generated video preserves the background and subject information from the conditioning image. 
All metrics are normalized scores, and higher values indicate better performance.

\paragraph{Imaging Quality.}
This metric evaluates low-level visual fidelity, such as sharpness and the absence of artifacts including blur, noise, and overexposure. 
VBench estimates frame-level image quality using an image quality predictor, such as MUSIQ, and averages the scores over all frames to obtain a video-level score.

\paragraph{Aesthetic Quality.}
This metric measures the visual appeal of generated frames, including realism, composition, and color harmony. 
VBench applies an aesthetic predictor, such as the LAION aesthetic model, to each frame and averages the resulting scores across the video.

\paragraph{Dynamic Degree.}
This metric measures the amount of motion in the generated video. 
Optical flow magnitudes, estimated for example by RAFT, are used to quantify motion intensity, encouraging generated clips to contain sufficiently dynamic rather than nearly static content.

\paragraph{Motion Smoothness.}
This metric evaluates whether the generated motion evolves smoothly over time and follows plausible temporal dynamics. 
VBench uses a pretrained video frame interpolation prior to assess the predictability and smoothness of intermediate motion, where more coherent motion receives a higher score.

\paragraph{Background Consistency.}
This metric evaluates the temporal stability of background appearance and layout. 
Frame-level visual features, such as CLIP features, are compared across time, and large temporal variations are penalized as background flickering or inconsistency.

\paragraph{Subject Consistency.}
This metric measures whether the foreground subject remains temporally consistent throughout the generated video. 
VBench compares subject-region features across frames to penalize identity drift, deformation, or abrupt appearance changes.

Formally, given the six VBench scores $\{s_k\}_{k=1}^{6}$ for a generated video, we define the overall score as the arithmetic mean of all dimensions:
\begin{equation}
    \mathrm{Overall\ Score}
    =
    \frac{1}{6}
    \sum_{k=1}^{6} s_k .
    \label{eq:vbench_i2v_overall}
\end{equation}
This averaged score is reported as the ``Overall Score'' in the main paper.

\subsection{4D Control Metrics}
We evaluate whether the generated video follows the intended 4D edit by measuring camera-motion accuracy and object-motion accuracy in the reconstructed 4D space. For each generated video, we estimate a camera trajectory and a set of dynamic object trajectories using the same 4D annotation protocol as used for constructing the benchmark. The target trajectory is obtained by applying the user-specified edit to the original reconstructed 4D scene. Both generated and target trajectories are expressed relative to their respective first frames before comparison.

For camera control, let $\{\mathbf{R}^{t}_{\mathrm{gt}}, \mathbf{T}^{t}_{\mathrm{gt}}\}_{t=1}^{T}$ denote the target camera rotations and translations, and let $\{\mathbf{R}^{t}_{\mathrm{gen}}, \mathbf{T}^{t}_{\mathrm{gen}}\}_{t=1}^{T}$ denote the camera trajectory estimated from the generated video. We measure rotation error by the geodesic distance on $\mathrm{SO}(3)$:
\begin{equation}
\mathrm{Cam\mbox{-}RotErr}
=
\frac{180}{\pi}\,\frac{1}{T}
\sum_{t=1}^{T}
\arccos
\left(
\frac{
\mathrm{tr}\!\left(
\mathbf{R}^{t}_{\mathrm{gen}}
{\mathbf{R}^{t}_{\mathrm{gt}}}^{\top}
\right)-1
}{2}
\right).
\end{equation}
The factor $180/\pi$ converts radians to degrees; Cam-RotErr is reported in degrees throughout. We measure translation error by the average Euclidean distance between the target and generated camera centers:
\begin{equation}
\mathrm{Cam\mbox{-}TransErr}
=
\frac{1}{T}
\sum_{t=1}^{T}
\left\|
\mathbf{T}^{t}_{\mathrm{gen}}
-
\mathbf{T}^{t}_{\mathrm{gt}}
\right\|_{2}.
\end{equation}
Lower values indicate better adherence to the intended camera motion.

For object motion control, we evaluate only the objects whose 4D trajectories are explicitly edited. Let $N_{\mathrm{gt}}$ be the number of controlled target objects and $N_{\mathrm{pred}}$ be the number of dynamic object trajectories estimated from the generated video. For target object $o$ and predicted object $k$, we define their trajectory distance as
\begin{equation}
d(o,k)
=
\frac{1}{T}
\sum_{t=1}^{T}
\left\|
\hat{\boldsymbol{\mu}}^{t}_{k}
-
\boldsymbol{\mu}^{t}_{o}
\right\|_{2},
\end{equation}
where $\boldsymbol{\mu}^{t}_{o}$ and $\hat{\boldsymbol{\mu}}^{t}_{k}$ are the 3D centers of the target and generated object trajectories at frame $t$. Since object identities are not known in the generated video, we perform bipartite matching between target and predicted trajectories using the Hungarian algorithm. Unmatched target objects are assigned a fixed penalty $\lambda$. The object motion control score is then
\begin{equation}
\mathrm{ObjMC}
=
\frac{1}{N_{\mathrm{gt}}}
\sum_{o=1}^{N_{\mathrm{gt}}}
d_o,
\quad
d_o =
\begin{cases}
d(o,k), & \text{if target object } o \text{ is matched to prediction } k,\\
\lambda, & \text{if } o \text{ is unmatched}.
\end{cases}
\end{equation}
This metric penalizes both inaccurate object trajectories and missing controlled objects. All three metrics are reported with lower values indicating better 4D control fidelity.

\section{Validation of PREBench Diagnostic Metrics}
\label{app:metric_validation}

We use two human-annotation stages to examine the proposed diagnostic metrics: individual-video severity ratings calibrate each metric's reference scale, and subsequent pairwise preferences test whether each metric orders outputs consistently with human judgments of its target artifact.

\paragraph{Study design.}
We sample 270 comparisons in total, with 30 for each diagnostic metric: 15 from the PREBench test split and 15 from an independently constructed calibration set randomly drawn from the SpatialVID-HQ and DynPose-100K dataset. For each metric, we construct eligible candidate pairs from outputs of different methods and ablations within each set, retaining only cases where the corresponding target region is non-empty. We compute the absolute metric gap for every eligible pair and partition the candidates into three equal-frequency gap bins: low, medium, and high. We randomly sample 30 pairs per metric while retaining 10 pairs from each gap bin and 15 from each set. The selected outputs are anonymized for individual-video severity ratings; the two videos in each pair are randomly ordered for the subsequent preference comparison.

\paragraph{Perceptual reference-scale calibration.}
Pairwise preferences establish an ordering but do not identify an absolute error level corresponding to a particular degree of degradation. We therefore supplemented the pairwise annotation pipeline with individual-video severity ratings to determine each half-score anchor from human judgments. The 30 pairs per metric yielded 60 unique rated outputs for each metric, with half drawn from each set (Table~\ref{tab:anchor_calibration_appendix}).

Before the pairwise comparison, each of the three annotators independently rated each anonymized video, with the source video, editing context, and target region available, but without method identities or metric values. The five severity levels were: 0, no observable target defect; 1, a slight defect noticeable on close inspection; 2, a clearly visible, moderate defect with the relevant content still largely intact and coherent; 3, a severe defect substantially disrupting appearance or coherence; and 4, an extreme defect causing the target region to fail. The instructions specified the target artifact for each metric: perceptual appearance change, structural distortion, temporal drift, and dynamic-content appearance change for the four Preserve metrics, respectively; residual ghosting and boundary discontinuity for Reveal; and temporal instability, boundary discontinuity, and copying/stretching for Expand.

For unique output $i$ and metric $m$, let $s_{i,m}$ be the median of the three severity ratings and $x_{i,m}$ its raw error. We determined the half-score anchor as
\begin{equation}
 h_m = \operatorname{median}_{i:\,s_{i,m}=2} x_{i,m}.
\label{eq:human_calibrated_anchor}
\end{equation}
This associates 50 points with the typical error among outputs judged to have moderate target degradation. Table~\ref{tab:anchor_calibration_appendix} reports the numbers of unique and moderate-rated outputs, the resulting anchors, and 95\% percentile bootstrap intervals from 2,000 resamples clustered by source case.

\begin{table}[t]
\centering
\caption{\textbf{Perceptual reference-scale calibration results.}
$N$ counts unique rated outputs, split equally between the PREBench test split and the independent calibration set; $N_2$ counts outputs with median severity 2 (moderate). The human-calibrated anchors $h_m$ were used to compute all reported aggregate PREBench scores. Intervals are 95\% percentile bootstrap intervals from 2,000 resamples clustered by source case.}
\label{tab:anchor_calibration_appendix}
\small
\begin{tabular}{lrrrc}
\toprule
Metric & $N$ & $N_2$ & $h_m$ & 95\% interval \\
\midrule
P-LPIPS     & 60 & 18 & 0.250 & [0.210, 0.300] \\
P-DISTS     & 60 & 16 & 0.150 & [0.120, 0.190] \\
P-TempDrift & 60 & 17 & 0.080 & [0.060, 0.105] \\
P-Dyn-LPIPS & 60 & 14 & 0.250 & [0.200, 0.310] \\
R-Ghost     & 60 & 19 & 0.200 & [0.160, 0.250] \\
R-Seam      & 60 & 15 & 0.080 & [0.060, 0.105] \\
E-Temp      & 60 & 18 & 0.030 & [0.022, 0.041] \\
E-Seam      & 60 & 16 & 0.080 & [0.060, 0.107] \\
E-Copy      & 60 & 20 & 0.300 & [0.250, 0.360] \\
\bottomrule
\end{tabular}
\end{table}

\paragraph{Pairwise preference annotation.}
After the individual-video ratings, each pair was evaluated by three independent annotators using a metric-specific two-alternative forced choice question. 
For R-Ghost, annotators are asked which video better completes the newly revealed region with fewer ghosting or residual artifacts. 
For R-Seam, annotators judge which video has a smoother transition between preserved and completed Reveal content. 
For E-Copy, annotators are asked which video better synthesizes expanded out-of-view content without copying, stretching, or repeating boundary content. 
For E-Seam, annotators judge which video has a smoother transition between preserved and newly generated regions. 
For E-Temp, annotators judge which video has more temporally stable expanded content. 
For P-LPIPS and P-DISTS, annotators judge source-backed perceptual appearance and structural detail, respectively; for P-TempDrift and P-Dyn-LPIPS, they judge temporal stability and dynamic-content appearance in Preserve regions, respectively. 
The human preference for each pair is determined by majority vote.

\paragraph{Evaluation protocol.}
All nine metrics are error-style metrics, where lower values indicate better results. 
Given a pair of outputs $(A, B)$ and a diagnostic metric $m$, the metric predicts $A$ to be preferred if $m(A) < m(B)$, and predicts $B$ otherwise. 
We measure the agreement between the metric prediction and the human majority preference:
\begin{equation}
\mathrm{Agreement}(m) =
\frac{1}{N_m}
\sum_{i=1}^{N_m}
\mathbf{1}
\left[
\mathrm{winner}_{m}^{(i)}
=
\mathrm{winner}_{human}^{(i)}
\right],
\end{equation}
where $N_m$ is the number of evaluated pairs for metric $m$.

We also report the correlation between metric margins and human vote margins. 
For each pair, we define the metric margin as
\begin{equation}
\Delta m = |m(A) - m(B)|,
\end{equation}
and the human vote margin as
\begin{equation}
\Delta h =
\left|
\frac{\#A}{\#A + \#B}
-
0.5
\right|,
\end{equation}
where $\#A$ and $\#B$ denote the number of annotators preferring outputs $A$ and $B$, respectively. 
We then compute Spearman's rank correlation between $\Delta m$ and $\Delta h$ over the evaluated pairs. 
With three annotators, $\Delta h$ takes only two values, corresponding to a 2--1 split or unanimous preference. The correlation therefore measures whether larger metric differences tend to coincide with unanimous votes, rather than a continuous degree of perceptual confidence.

\begin{table}[t]
\centering
\caption{\textbf{Validation of PREBench metrics against human preference.}
We evaluate whether each of the nine PREBench diagnostics predicts the human majority preference in two-alternative forced choice comparisons, with 15 pairs per metric from each set. 
Agreement measures how often the metric-predicted winner matches the human-preferred winner. 
Spearman's $\rho$ measures the correlation between metric margin and vote unanimity, encoded by the two possible human vote margins.}
\label{tab:metric_validation_appendix}
\resizebox{0.86\linewidth}{!}{
\begin{tabular}{lcccc}
\toprule
Metric 
& Evaluated artifact 
& \#Pairs 
& Agreement $\uparrow$ 
& Spearman $\rho$ $\uparrow$ \\
\midrule
P-LPIPS 
& Preserve perceptual similarity 
& 30 
& 83.3\% 
& 0.68 \\
P-DISTS 
& Preserve structural similarity 
& 30 
& 80.0\% 
& 0.61 \\
P-TempDrift 
& Preserve temporal drift 
& 30 
& 76.7\% 
& 0.55 \\
P-Dyn-LPIPS 
& Preserve dynamic-content fidelity 
& 30 
& 73.3\% 
& 0.49 \\
R-Ghost 
& Ghosting in Reveal regions 
& 30 
& 76.7\% 
& 0.52 \\
R-Seam 
& Boundary discontinuity in Reveal regions 
& 30 
& 70.0\% 
& 0.41 \\
E-Temp 
& Temporal instability in Expand regions 
& 30 
& 76.7\% 
& 0.49 \\
E-Seam 
& Boundary discontinuity in Expand regions 
& 30 
& 66.7\% 
& 0.38 \\
E-Copy 
& Copying/stretching in Expand regions 
& 30 
& 80.0\% 
& 0.53 \\
\midrule
Average 
& -- 
& 270 
& 75.9\% 
& 0.52 \\
\bottomrule
\end{tabular}
}
\end{table}

\begin{figure}[t]
  \centering
  \includegraphics[width=\linewidth]{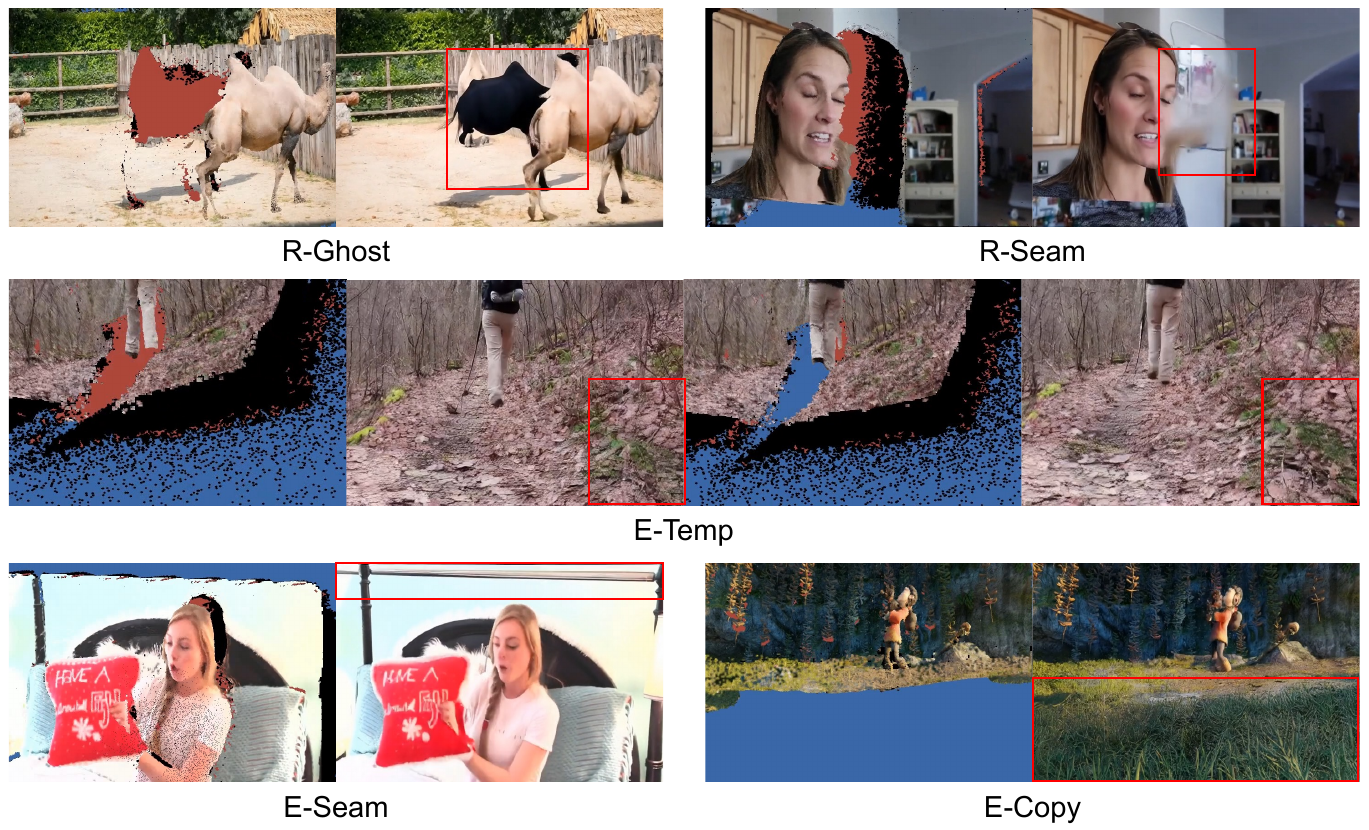}
  \caption{\textbf{Failure modes captured by PREBench diagnostic metrics.}}
  \label{fig:prex_failure_cases}
\end{figure}

\paragraph{Discussion.}
Across the stratified comparison sample, the nine evaluated diagnostics agree with human majority preference in 66.7--83.3\% of pairs (Table~\ref{tab:metric_validation_appendix}), supporting their utility for distinguishing the corresponding region-specific artifacts in this sample. 
The strongest agreement is obtained by P-LPIPS (83.3\%), followed by P-DISTS and E-Copy (80.0\% each), indicating that both Preserve fidelity and degenerate Expand behavior are comparatively well captured by the corresponding metrics. 
The remaining Preserve metrics also align well with human judgments: P-TempDrift and P-Dyn-LPIPS reach 76.7\% and 73.3\% agreement, respectively, supporting their use for identifying temporal drift and loss of dynamic content in preserved regions. 
The rank correlations show the same overall pattern: P-LPIPS has the strongest association with vote unanimity (\(\rho=0.68\)), followed by P-DISTS (0.61) and E-Copy (0.53); all nine correlations are positive. 
While these metrics do not fully replace human judgment, they aim to identify potential region-specific failures at scale and reduce the need for labor-intensive human evaluation in every comparison.

\paragraph{Visualization.}

We visualize typical failures diagnosed by PREBench metrics in Fig~\ref{fig:prex_failure_cases}. The examples illustrate that different PREBench metrics capture complementary failure modes. 
R-Ghost highlights cases where invalid evidence from the removed or occluded object remains visible in the generated reveal region, while R-Seam captures unnatural transitions between completed reveal content and preserved source-backed areas. 
For expanded views, E-Temp identifies temporally unstable synthesis in newly generated out-of-view regions, and E-Seam reflects visible boundary discontinuities between expanded and preserved regions. 
E-Copy further exposes degenerate expansion behavior, where the model copies, stretches, or repeats textures from the original field-of-view boundary instead of synthesizing plausible new content. 
These qualitative examples support the diagnostic role of PREBench: while the aggregate score enables method-level ranking, the component metrics localize distinct region-specific artifacts and make the failure modes of 4D video editing methods more interpretable.

\subsection{Separate Overall-Preference Study for PREBench Score}
\label{app:prebench_overall_preference}

We conducted a separate two-alternative forced-choice (2AFC) study to assess whether PREBench Score agrees with overall human preference. Using random seed 0, we sampled 30 distinct source-video cases from the independent calibration set and paired outputs of two different methods for the same source video and target edit, yielding 30 pairs and 60 outputs. Five independent annotators judged each pair. Pairs were sampled without reference to metric values or PREBench Scores.

Annotators viewed the source video, the requested edit, and two anonymized outputs in randomized order. They selected the output that better preserved source-backed content while producing plausible, temporally coherent content in newly revealed and out-of-view regions. Method identities and scores were hidden. The output with the higher PREBench Score matched the human majority preference in 26 of 30 pairs, giving 86.7\% agreement. We report a two-sided 95\% Wilson score interval for this proportion in Table~\ref{tab:prebench_overall_preference_appendix}.

\begin{table}[t]
\centering
\caption{\textbf{Separate overall 2AFC validation of PREBench Score.}}
\label{tab:prebench_overall_preference_appendix}
\small
\resizebox{\linewidth}{!}{%
\begin{tabular}{lccccc}
\toprule
Study & Source cases & Rated outputs & Pairs & Agreement $\uparrow$ & 95\% interval \\
\midrule
Overall preference & 30 & 60 & 30 & 86.7\% & [70.3\%, 94.7\%] \\
\bottomrule
\end{tabular}
}
\end{table}

\section{Controlled Evaluation and Robustness}
\label{app:evaluation_isolation}

We use three controlled evaluations to examine conditioning contributions and sensitivity to mask alignment, observation reuse, and upstream reconstruction quality. The evaluation masks are fixed for each PREBench case and applied identically to every method.

\subsection{Conditioning-mask Perturbation}

To explicitly decouple conditioning masks from evaluation masks, we perturb only the role-mask channels at inference, without retraining or changing the RGB cue. We consider a fixed 20-pixel spatial shift and random reassignment of 10\% of role labels. The original, unperturbed PREBench masks are retained exclusively for evaluation. Table~\ref{tab:mask_perturbation_appendix} reports the complete set of global and region-aware diagnostics.

\begin{table}[t]
\centering
\caption{\textbf{Robustness to perturbed conditioning masks.} Evaluation always uses the original unperturbed annotations.}
\label{tab:mask_perturbation_appendix}
\scriptsize
\setlength{\tabcolsep}{3.2pt}
\resizebox{\linewidth}{!}{
\begin{tabular}{lccccccccccc}
\toprule
Conditioning mask & VBench $\uparrow$ & P-LPIPS $\downarrow$ & P-DISTS $\downarrow$ & P-TempDrift $\downarrow$ & P-Dyn-LPIPS $\downarrow$ & R-Ghost $\downarrow$ & R-Seam $\downarrow$ & E-Temp $\downarrow$ & E-Seam $\downarrow$ & E-Copy $\downarrow$ & \begin{tabular}[c]{@{}c@{}}PREBench\\Score $\uparrow$\end{tabular} \\
\midrule
Clean & 77.99 & 0.2137 & 0.1199 & 0.0532 & 0.0319 & 0.1374 & 0.0277 & 0.0638 & 0.0308 & 0.3764 & \textbf{61.45} \\
20-pixel spatial shift & 77.69 & 0.2150 & 0.1263 & 0.0585 & 0.0340 & 0.1686 & 0.0323 & 0.0871 & 0.0485 & 0.3838 & 57.09 \\
10\% random corruption & 77.82 & 0.2176 & 0.1241 & 0.0548 & 0.0327 & 0.1486 & 0.0286 & 0.0672 & 0.0406 & 0.3799 & \underline{59.80} \\
\bottomrule
\end{tabular}
}
\end{table}

Global quality remains stable under both perturbations, whereas region diagnostics degrade in a structured manner: PREBench Score decreases from 61.45 to 57.09 under the spatial shift and to 59.80 under random corruption, corresponding to drops of 4.36 and 1.65 points. The spatial shift has the largest effect on R-Ghost and boundary-sensitive errors because it systematically misaligns synthesis roles with edit boundaries. Random corruption is milder because most labels and local boundaries remain correct.

\subsection{Disjoint-observation Preserve Evaluation}

We further evaluate preservation under a five-fold disjoint-observation protocol that separates cue construction from reference evaluation. In each fold, every fifth eligible source frame is reserved exclusively for evaluation, and its immediate temporal neighbors are excluded as a buffer. The remaining frames construct the appearance cue. Across five folds, every eligible source frame is evaluated once. The mandatory first-frame I2V reference is retained for all methods but is never used as a held-out evaluation observation. Held-out Preserve references are reconstructed only from reserved frames using the same visibility, depth, instance, and projection checks. We additionally report P-SelfTemp, an output-only temporal-consistency diagnostic that uses no source RGB reference.

\begin{table}[t]
\centering
\caption{\textbf{Disjoint-observation evaluation in Preserve regions.} Evaluation observations and their immediate neighbors are excluded from cue construction.}
\label{tab:heldout_preserve_appendix}
\scriptsize
\setlength{\tabcolsep}{4.0pt}
\resizebox{0.88\linewidth}{!}{
\begin{tabular}{lccccc}
\toprule
Method & Held-out P-LPIPS & Held-out P-DISTS & Held-out P-TempDrift & Held-out P-Dyn-LPIPS & P-SelfTemp \\
\midrule
GEN3C & 0.3753 & 0.1845 & 0.0878 & 0.0612 & 0.0692 \\
Plain Wan2.1-ControlNet & 0.4124 & 0.1874 & 0.0992 & 0.1029 & 0.0813 \\
PREX w/o o-b appearance cues & 0.4273 & 0.1826 & 0.0865 & 0.0677 & 0.0741 \\
Full PREX & \textbf{0.2712} & \textbf{0.1487} & \textbf{0.0571} & \textbf{0.0342} & \textbf{0.0589} \\
\bottomrule
\end{tabular}
}
\end{table}

Full PREX achieves the lowest errors on all held-out Preserve metrics and the lowest output-only P-SelfTemp. Relative to PREX without observation-backed cues, it reduces held-out P-LPIPS, P-DISTS, P-TempDrift, and P-Dyn-LPIPS by 36.5\%, 18.6\%, 34.0\%, and 49.5\%, respectively.

\subsection{Sensitivity to Upstream 4D Errors}

We independently perturb geometry and visibility at inference without retraining. Moderate and severe geometry noise add zero-mean Gaussian offsets with standard deviations equal to 1\% and 2\% of the scene scale. Visibility perturbations randomly flip 10\% or 20\% of visibility labels near occlusion boundaries. The appearance cues, confidence maps, and conditioning role masks are regenerated from the perturbed proxy, while the original PREBench masks remain fixed for evaluation.

\begin{table}[t]
\centering
\caption{\textbf{Sensitivity to upstream 4D errors.} Geometry and visibility are perturbed independently at inference.}
\label{tab:upstream_robustness_appendix}
\scriptsize
\setlength{\tabcolsep}{3.2pt}
\resizebox{\linewidth}{!}{
\begin{tabular}{llccccccccccc}
\toprule
Perturbation & Level & VBench $\uparrow$ & P-LPIPS $\downarrow$ & P-DISTS $\downarrow$ & P-TempDrift $\downarrow$ & P-Dyn-LPIPS $\downarrow$ & R-Ghost $\downarrow$ & R-Seam $\downarrow$ & E-Temp $\downarrow$ & E-Seam $\downarrow$ & E-Copy $\downarrow$ & \begin{tabular}[c]{@{}c@{}}PREBench\\Score $\uparrow$\end{tabular} \\
\midrule
None & Clean & 77.99 & 0.2137 & 0.1199 & 0.0532 & 0.0319 & 0.1374 & 0.0277 & 0.0638 & 0.0308 & 0.3764 & \textbf{61.45} \\
Geometry & Moderate & 75.62 & 0.2510 & 0.1390 & 0.0638 & 0.0397 & 0.1530 & 0.0331 & 0.0789 & 0.0387 & 0.3689 & 57.60 \\
Geometry & Severe & 71.84 & 0.3260 & 0.1740 & 0.0825 & 0.0548 & 0.1840 & 0.0428 & 0.1091 & 0.0529 & 0.3767 & 51.22 \\
Visibility & Moderate & 76.71 & 0.2260 & 0.1270 & 0.0571 & 0.0348 & 0.1540 & 0.0320 & 0.0698 & 0.0346 & 0.3856 & \underline{59.15} \\
Visibility & Severe & 74.35 & 0.2490 & 0.1380 & 0.0640 & 0.0395 & 0.1810 & 0.0391 & 0.0826 & 0.0418 & 0.3712 & 55.71 \\
\bottomrule
\end{tabular}
}
\end{table}

Moderate geometry and visibility perturbations reduce PREBench Score by 3.85 and 2.30 points, respectively. At the tested perturbation levels, geometry noise causes larger score reductions, consistent with its effects on projected appearance, depth consistency, region boundaries, and temporal correspondence. Visibility errors are more localized and primarily increase ghost leakage and boundary artifacts by exposing invalid evidence or misclassifying disocclusions. These results quantify the sensitivity of PREX to upstream 4D quality.

\section{Additional Comparison and Baseline Protocol}
\label{app:baseline_protocol}

\subsection{Baseline Adaptation and Tuning Protocol}

Within each reported table, all methods are evaluated on the same supported cases and at the same input resolution. We express each intended edit through the method's native interface, while retaining the released checkpoint and recommended inference configuration whenever available. No released baseline is fine-tuned for PREBench or selected using its region-aware metrics; method-specific changes are limited to the control-format conversion summarized in Table~\ref{tab:baseline_adaptation}.

\begin{center}
\refstepcounter{table}
\label{tab:baseline_adaptation}
\vspace{0.5em}
\parbox{\textwidth}{\centering Table~\thetable: \textbf{Baseline adaptation protocol.} The same edited 4D proxy and target controls are converted to each method's native interface.}
\vspace{0.5em}
\centering
\scriptsize
\setlength{\tabcolsep}{1.3pt}
\renewcommand{\arraystretch}{1.08}
\begin{tabular}{p{0.11\textwidth}p{0.16\textwidth}p{0.15\textwidth}p{0.34\textwidth}p{0.20\textwidth}}
\toprule
Method & Base model & Native input & Conversion from the common edit & Checkpoint and configuration \\
\midrule
Perception-as-Control
& Stable Diffusion 1.5
& 3D-aware motion representation
& Convert the target camera and object motion from the edited proxy into the method's native motion-control representation.
& Released checkpoint and recommended inference settings. \\
Yume
& Yume-MVDT
& Interactive camera and scene control
& Convert the target camera trajectory and available scene-motion controls into the supported interface.
& Released checkpoint and default inference configuration. \\
DaS
& CogVideoX-5B-I2V
& Rendered 3D-aware geometric conditions
& Render the edited proxy in the geometric condition format required by DaS using the same target cameras.
& Released checkpoint and recommended conditioning strength. \\
GEN3C
& Cosmos-7B Video2World
& 3D-informed rendered guidance and camera control
& Render the edited proxy with the target cameras and convert it into GEN3C's native guidance representation.
& Released checkpoint and official inference configuration. \\
VerseCrafter
& Wan2.1-T2V-14B
& 4D geometric control
& Convert proxy geometry, camera trajectory, and object motion into the method's native 4D control representation.
& Released checkpoint and recommended settings. \\
FlashWorld
& Wan2.2-5B-IT2V
& 3D scene guidance and camera control
& Convert the proxy into the supported 3D scene condition and apply the same target camera trajectory on compatible cases.
& Released checkpoint and default settings. \\
Voyager
& HunyuanVideo-I2V
& 3D scene representation and camera trajectory
& Convert the edited proxy into the supported scene representation and apply the target camera trajectory on compatible camera edits.
& Released checkpoint and default generation configuration. \\
NeoVerse
& Wan2.1-T2V-14B
& Reconstructed 4D scene and reshooting controls
& Convert the edited scene and target cameras into the native 4D reshooting representation on camera-only cases.
& Released checkpoint and recommended inference settings; no object-editing extension. \\
Vista4D
& Wan2.1-T2V-14B
& 4D point-cloud guidance and target cameras
& Convert the edited proxy into Vista4D-compatible point-cloud guidance and evaluate it on the camera-reshooting subset.
& Official checkpoint and recommended inference configuration. \\
Plain Wan2.1-ControlNet
& Wan2.1-I2V-14B
& Raw RGB control sequence
& Directly render the edited proxy into raw RGB control frames.
& Newly trained matched control with the same backbone, data, resolution, training budget, and inference settings as PREX. \\
\bottomrule
\end{tabular}
\end{center}

\begin{center}
\refstepcounter{table}
\label{tab:conventional_editing_comparison}
\vspace{0.5em}
\parbox{\textwidth}{\centering Table~\thetable: \textbf{Qualitative comparison with general-purpose video editing methods.}}
\vspace{0.5em}
\scriptsize
\setlength{\tabcolsep}{3pt}
\renewcommand{\arraystretch}{1.08}
\begin{tabular}{p{0.13\textwidth}p{0.18\textwidth}p{0.28\textwidth}p{0.35\textwidth}}
\toprule
Method & Primary task and control & Characteristic strength & Relation to PREX scope \\
\midrule
FateZero~\citep{qi2023fatezero}
& Text-driven appearance and attribute editing
& Zero-shot attention fusion preserves temporal consistency and source structure during semantic edits.
& Operates on semantic appearance within the original image-plane layout and motion; PREX targets explicit 4D transformations and the resulting visibility changes. \\
TokenFlow~\citep{geyer2023tokenflow}
& Text-guided video editing with propagated diffusion features
& Preserves the source video's spatial layout and dynamics while enforcing cross-frame feature consistency without task-specific training.
& Propagates appearance edits through existing correspondences; PREX edits the underlying 4D scene configuration. \\
Rerender-A-Video~\citep{yang2023rerender}
& Text-guided video-to-video translation
& Uses coherent key-frame translation and patch propagation for consistent shape, texture, and color editing.
& Targets rerendering and stylization of observed content, while PREX uses explicit 4D camera/object controls and evidence roles. \\
VidToMe~\citep{li2024vidtome}
& Zero-shot appearance editing through cross-frame token merging
& Improves short- and long-range temporal consistency while reducing attention memory.
& Extends image editing to video while largely retaining input motion; PREX models unsupported regions created by 4D edits. \\
VideoDirector~\citep{wang2025videodirector}
& Text-driven precise video editing
& Constrains reverse diffusion with source reconstruction to balance editability and fidelity.
& Changes semantic appearance under the source-video structure; PREX executes camera and object transformations in a reconstructed 4D scene. \\
ProPainter~\citep{zhou2023propainter}
& Mask-based video inpainting and object removal
& Propagates reliable content over long temporal ranges and fills prescribed missing regions coherently.
& Provides 2D masked completion; PREX additionally controls target camera/object motion and separates in-scene Reveal from out-of-view Expand regions. \\
VACE~\citep{jiang2025vace}
& Unified V2V, masked V2V, and reference-conditioned creation
& Composes video, mask, image, and text conditions through a unified interface for diverse editing and generation tasks.
& Provides a broad 2D editing interface; PREX focuses on explicit 4D scene transformations with Preserve, Reveal, and Expand roles. \\
\bottomrule
\end{tabular}
\end{center}

\subsection{Qualitative Comparison with Conventional Video Editing}
\label{app:conventional_editing_comparison}

Tables~\ref{tab:conventional_editing_comparison} and~\ref{tab:motion_editing_comparison} summarize task scope, control interfaces, and treatment of newly unsupported content for representative methods, based on their published formulations.

\begin{center}
\refstepcounter{table}
\label{tab:motion_editing_comparison}
\vspace{0.5em}
\parbox{\textwidth}{\centering Table~\thetable: \textbf{Qualitative comparison with motion- and camera-controlled video methods.}}
\vspace{0.5em}
\scriptsize
\setlength{\tabcolsep}{3pt}
\renewcommand{\arraystretch}{1.08}
\begin{tabular}{p{0.13\textwidth}p{0.18\textwidth}p{0.28\textwidth}p{0.35\textwidth}}
\toprule
Method & Primary task and control & Characteristic strength & Relation to PREX scope \\
\midrule
MotionCtrl~\citep{wang2024motionctrl}
& Text-to-video generation with camera poses and object trajectories
& Provides unified, independently composable camera- and object-motion control.
& Generates from text and motion conditions; PREX edits an existing source video and preserves valid observations. \\
DragAnything~\citep{wu2024draganything}
& Image-to-video object motion control with user trajectories
& Uses entity-level representations to move arbitrary selected objects while retaining their visual identity.
& Uses a reference image and 2D motion input; PREX uses full-video observations together with target-view visibility and scene extent. \\
Wan-Move~\citep{chu2025wan}
& Image-to-video generation with dense point trajectories
& Propagates first-frame latent features along trajectories for scalable fine-grained motion control and motion transfer.
& Anchors generation to the first frame; PREX verifies evidence across the source video and separates disocclusion from out-of-view expansion. \\
MotionV2V~\citep{burgert2026motionv2v}
& Video-to-video motion editing with source and target 2D point trajectories
& Edits object or camera motion from arbitrary timestamps while conditioning on the full input video for content preservation.
& Provides a flexible sparse-track interface; PREX uses reconstructed scene extent to partition source-supported, disoccluded, and out-of-view regions. \\
Edit-by-Track~\citep{lee2026generative}
& Joint camera/object motion editing with editable 3D point tracks
& Uses source/target projected 3D tracks as sparse correspondences, supporting camera motion, object motion, deformation, removal, and duplication.
& Uses sparse 3D track control; PREX derives appearance cues and evidence roles from an edited 4D proxy. \\
PREX (Ours)
& 4D-driven camera and object editing with an edited 4D proxy
& Uses observation-backed cues, confidence, and explicit evidence roles to preserve valid content and synthesize unsupported regions selectively.
& Focuses on faithful 4D reshooting and motion editing with explicit observation support and synthesis roles. \\
\bottomrule
\end{tabular}
\end{center}

Appearance-oriented editors are particularly suitable for changing style, identity, or local attributes while retaining the original motion and layout. ProPainter supports user-masked object removal and video completion, while VACE unifies a broader collection of video, mask, and reference-conditioned tasks. MotionCtrl, DragAnything, and Wan-Move provide motion control for generated videos, principally from text or a reference image. MotionV2V and Edit-by-Track illustrate trajectory-based editing of existing videos, using source and target 2D tracks and editable 3D point tracks, respectively. PREX instead uses an edited dense 4D proxy to retrieve source-supported appearance and distinguish in-scene Reveal completion from out-of-view Expand synthesis. At the time of evaluation, public inference implementations or online APIs for running MotionV2V and Edit-by-Track on PREBench were unavailable, so these methods are included in the interface comparison but not the quantitative tables.

\begin{figure}[t]
  \centering
  \includegraphics[width=\linewidth]{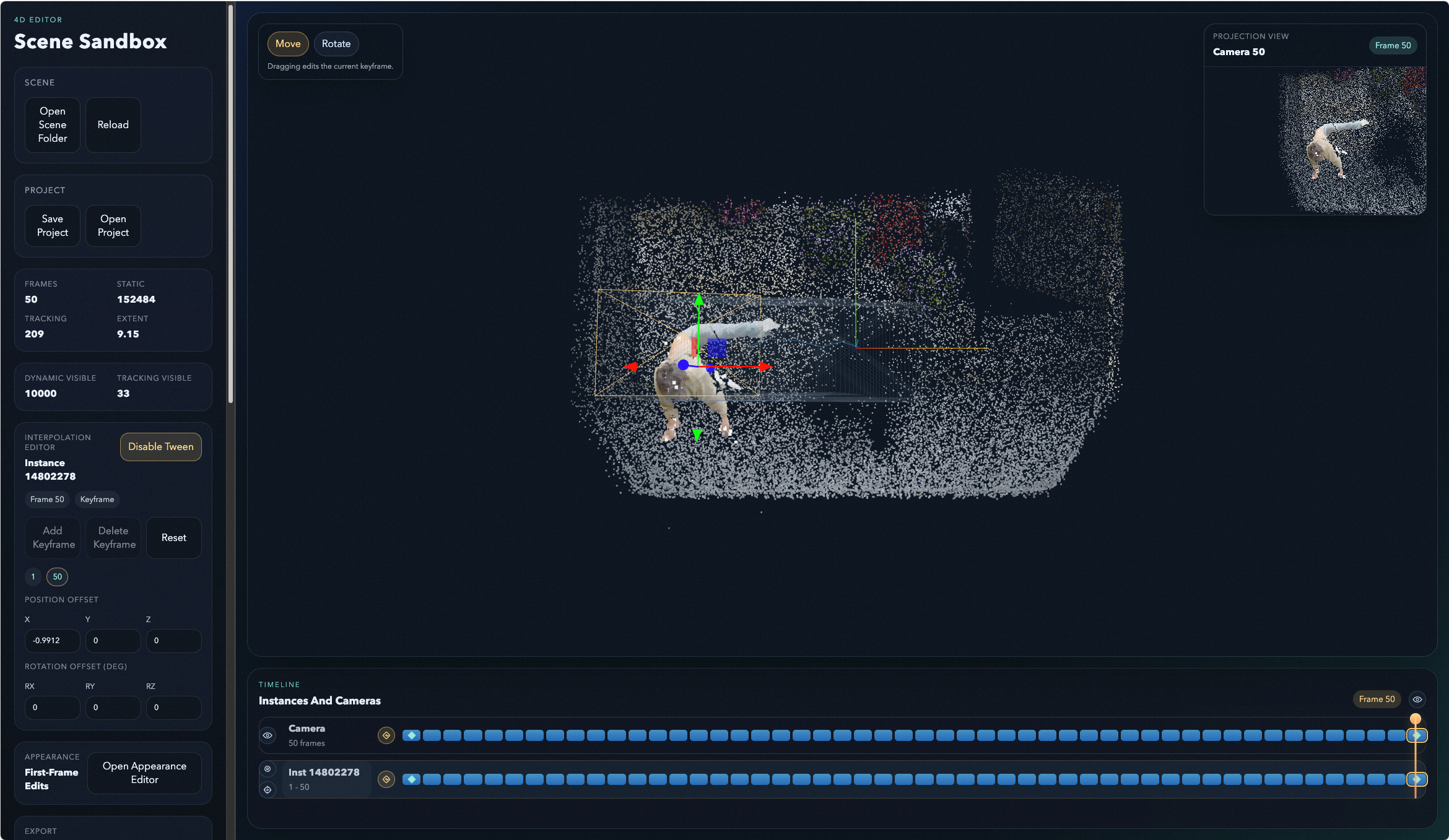}
  \caption{\textbf{Interactive 4D editing interface.}
  The interface provides a scene-level sandbox for editing reconstructed 4D scenes. Users can load a scene, inspect the reconstructed point-based 4D representation, manipulate dynamic instances with interactive transform controls, preview the edited result from target camera views, and manage camera/object keyframes through a timeline.}
  \label{fig:gui}
\end{figure}

\section{Interactive General User Interface for 4D Editing}
\label{app:gui}

To facilitate practical 4D video editing, we implement an interactive graphical user interface for scene inspection, instance manipulation, and camera control, as shown in Fig.~\ref{fig:gui}. The interface visualizes the reconstructed 4D scene as a point-based representation and allows users to directly manipulate dynamic instances using translation and rotation controls. The left panel provides scene and project management functions, together with statistics such as the number of frames, tracked points, and visible dynamic instances. The central viewport supports interactive editing in the global 4D scene, while the projection window previews the current target camera view. At the bottom, a timeline displays both camera and instance tracks, enabling users to add, delete, and adjust keyframes for temporal editing. This interface allows users to specify object motion, camera reshooting, and appearance-related edits in an intuitive manner, producing the edited 4D proxy and target camera trajectories used by PREX for faithful video generation.

\section{Model Architecture Details}
\label{app:model_architecture}

PREX is implemented on top of the Wan2.1-I2V-14B backbone, a
latent video diffusion model with a 3D VAE and a DiT-based video denoiser.
We keep the pretrained backbone frozen and introduce a region-aware conditioning branch, while the first-frame reference and text
conditioning follow the original Wan-I2V pipeline.

For each target video, PREX provides three region-aware control signals:
the observation-backed RGB control \(C^{rgb}\), the confidence map
\(C^{conf}\), and the edit-region masks \(M^{R}, M^{E}\) for Reveal and
Expand regions. The RGB control is encoded by the frozen Wan VAE into a
latent feature \(z^{rgb}\). The confidence map is resized to the latent
resolution, and the edit masks are rearranged according to the VAE spatial
stride so that region labels remain aligned with the latent grid. In our setting, these signals are concatenated as
\[
G =
\mathrm{Concat}
\left(
z^{rgb},
\phi_{\mathrm{conf}}(C^{conf}),
\phi_{\mathrm{mask}}(M^{R}, M^{E})
\right),
\]
where \(G\) is the input to the Region Adapter branch. With the Wan VAE
compression ratios \(s_t=4\) and \(s_h=s_w=8\), the mask embedding folds
the two edit-region masks into latent-aligned spatial sub-cells, producing
64 mask channels. Together with 16 RGB latent channels and one confidence
channel, the Region Adapter input dimension is 81.

The Region Adapter branch uses the same token grid as the Wan-DiT video
tokens. The latent control tensor \(G\) is patchified and projected into
geometry-aware control tokens, which are injected into selected Wan-DiT
blocks as residual hints. In the 14B model, we inject the adapter at layers
\[
\mathcal{L}_{\mathrm{ada}} =
\{0,5,10,15,20,25,30,35\}.
\]
For a frozen Wan-DiT block \(B_{\ell}\), the update can be written as
\[
x_{\ell+1}
=
B_{\ell}(x_{\ell})
+
\mathbf{1}[\ell\in\mathcal{L}_{\mathrm{ada}}]\,
\alpha\, H_{m(\ell)}(G),
\]
where \(H_{m(\ell)}\) denotes the corresponding Region Adapter block and
\(\alpha\) is the conditioning strength. In our experiments, we set
\(\alpha=1.0\). Each Region Adapter block is initialized from its paired
Wan-DiT block, while the residual projections are initialized with a small
near-zero scale. Therefore, the model starts close to the original
pretrained video generator and gradually learns to use the PREX
region-aware controls for preservation, reveal completion, and expansion.

We train with the AdamW optimizer using bf16 mixed precision. The per-device batch size is set to 2 with gradient accumulation of 1, giving an effective batch size of 32. In the initial training stage, we train the PREX Region-Aware Adapter for 50 epochs with a learning rate of $2 \times 10^{-5}$ and 100 warmup steps, while keeping the pretrained backbone frozen. For proxy-task curriculum training, we resume from this checkpoint and train the adapter for another 50 epochs with a learning rate of $5 \times 10^{-6}$ and 50 warmup steps. We use a fixed random seed of 42 for reproducibility; in distributed training, each process uses a rank-offset seed, and the data sampler is re-seeded by epoch.

\section{Training Datasets and PREBench Test Set}
\label{app:datasets}

\subsection{Training Datasets}

The training data contain 10,000 training pairs collected from seven public video datasets, as summarized in Table~\ref{tab:training-data-composition}. A pair consists of one conditioning control sequence and its reconstruction target; the same source clip may appear under different condition types, but source-level splitting is performed before pair construction. Thus, clips and editing variants derived from one source video do not cross the training, validation, and PREBench test splits.
The final composition contains 4,589 DynPose-100K pairs, 4,590 SpatialVID-HQ pairs, and 821 pairs from UVO, PointOdyssey, Dynamic Replica, DAVIS, and Spring. This mixture combines dynamic scenes with rich camera and object motion from DynPose-100K and additional video coverage from SpatialVID-HQ. 
This mixed composition supports both reconstruction learning and proxy-task curriculum learning, where the model learns to preserve source-backed regions, complete newly revealed areas, and extrapolate out-of-view content under different motion and scene configurations.

\begin{table}[t]
  \centering
  \begin{minipage}[c]{0.34\textwidth}
    \centering
    \captionof{table}{Composition of Training Data.}
    \label{tab:training-data-composition}
    \resizebox{\linewidth}{!}{
    \begin{tabular}{lrr}
      \toprule
      Dataset & Pairs & Ratio (\%) \\
      \midrule
      DynPose-100K & 4{,}589 & 45.89 \\
      SpatialVID-HQ & 4{,}590 & 45.90 \\
      UVO & 345 & 3.45 \\
      PointOdyssey & 215 & 2.15 \\
      Dynamic Replica & 145 & 1.45 \\
      DAVIS & 111 & 1.11 \\
      Spring & 5 & 0.05 \\
      \midrule
      Total & 10{,}000 & 100.00 \\
      \bottomrule
    \end{tabular}
    }
  \end{minipage}
  \hfill
    \begin{minipage}[c]{0.62\textwidth}
      \centering
      \captionof{table}{Composition of PREBench Test Set.}
      \label{tab:prebench-test-composition}
      \resizebox{\linewidth}{!}{
      \begin{tabular}{lrrrrrr}
        \toprule
        Dataset
        & \multicolumn{2}{c}{Camera-only}
        & \multicolumn{2}{c}{Camera+Object}
        & \multicolumn{2}{c}{Total} \\
        \cmidrule(lr){2-3}
        \cmidrule(lr){4-5}
        \cmidrule(lr){6-7}
        & Cases & Ratio (\%)
        & Cases & Ratio (\%)
        & Cases & Ratio (\%) \\
        \midrule
        DAVIS        & 76 & 50.67 & 76  & 38.00 & 152 & 43.43 \\
        DynPose-100K & 44 & 29.33 & 100 & 50.00 & 144 & 41.14 \\
        Spring       & 30 & 20.00 & 24  & 12.00 & 54  & 15.43 \\
        \midrule
        Total        & 150 & 100.00 & 200 & 100.00 & 350 & 100.00 \\
        \bottomrule
      \end{tabular}
      }
    \end{minipage}
\end{table}

\subsection{PREBench Test Set}

The PREBench test set is designed to evaluate realistic 4D editing scenarios under both camera-only and joint camera-object control settings. 
As shown in Table~\ref{tab:prebench-test-composition}, the PREBench test set contains 350 cases in total, including 150 camera-only cases and 200 camera-object cases. 
The camera-only subset evaluates whether a method can follow edited target camera trajectories while preserving valid source observations and synthesizing newly visible regions. 
The camera-object subset is more challenging, as object edits can introduce disocclusion and reveal regions while camera motion can simultaneously create out-of-view expansion regions. 
The test cases are drawn from DAVIS, DynPose-100K, and Spring, covering real videos and photo-realistic synthetic sequences with dynamic object motions and camera trajectories. 
For every test-set video, the algorithmically generated Preserve, Reveal, and Expand masks were manually checked and refined in a second annotation pass before evaluation. These verified region annotations remain fixed across all compared methods. 
This design enables PREBench to diagnose preservation fidelity, reveal-region completion, and expansion quality under practical 4D video editing operations.

\begin{figure*}[t]
  \centering
  \includegraphics[width=\textwidth]{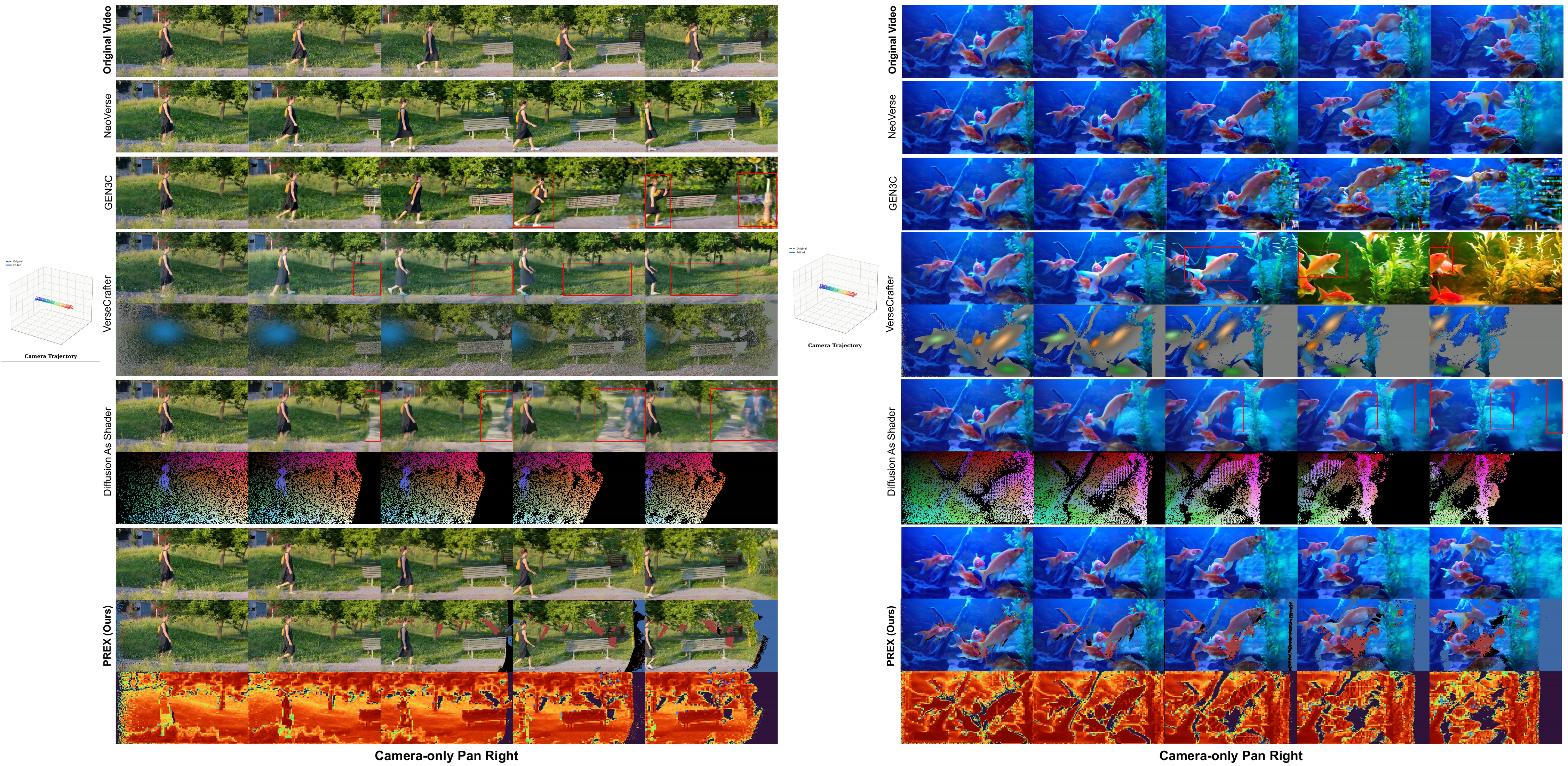}
  \caption{\textbf{More qualitative comparison of camera-only motion control on PREBench.}}
  \label{fig:qualitative_main}
\end{figure*}

\begin{figure*}[t]
  \centering
  \includegraphics[width=\textwidth]{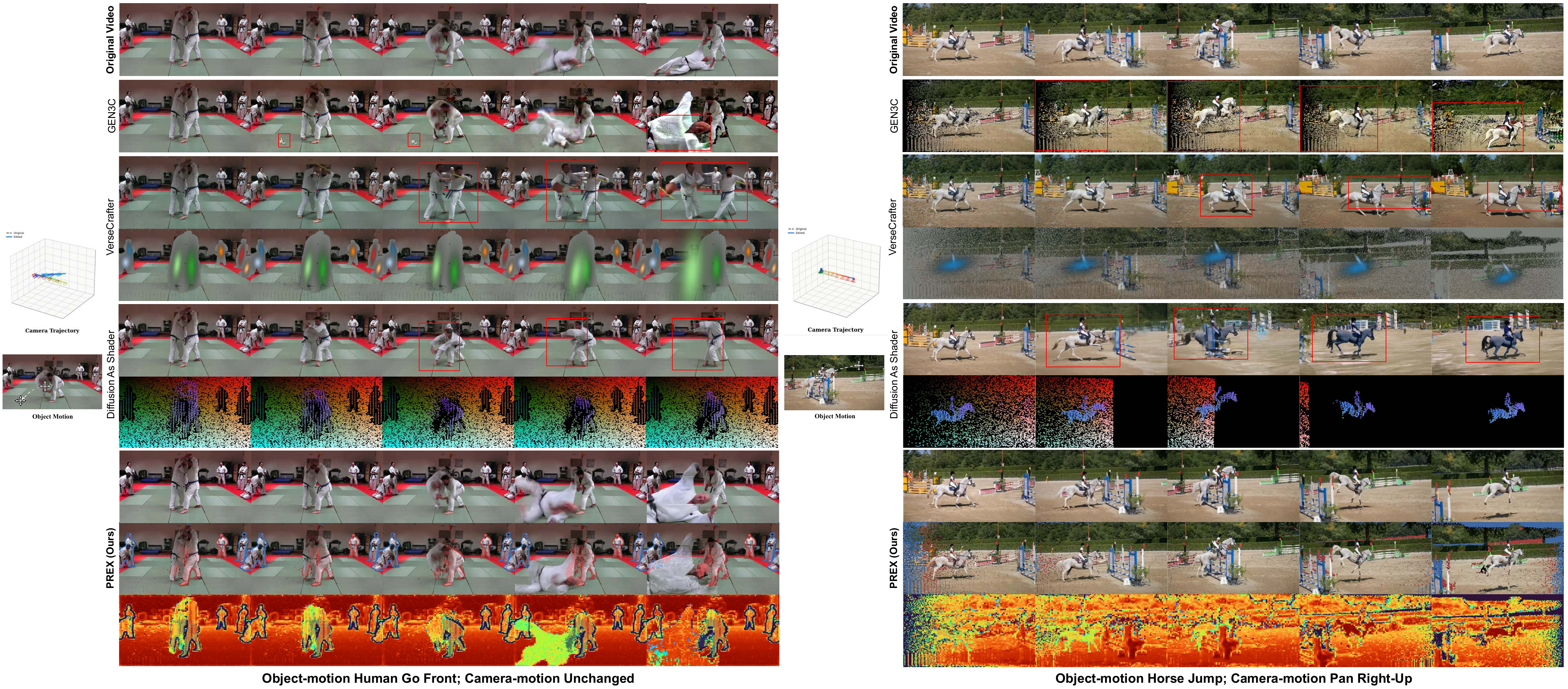}
  \caption{\textbf{More qualitative comparison of camera-object joint motion control on PREBench.}}
\end{figure*}

\section{Inference Configuration and Runtime}
\label{app:inference_runtime}

For inference, PREX uses a 50-step denoising schedule to generate each video. Unless otherwise specified, we use PREX with the Wan2.1-I2V-14B backbone and keep both the backbone and the Region-Aware Adapter fully resident in GPU memory during generation. Under this setting, a single PREX inference requires approximately 66GB of additional GPU memory. The end-to-end runtime for a cold-start single-video run is about 6 minutes, including model loading and initialization. Once the model has been loaded, the main generation stage for one 49-frame video takes approximately 4 minutes. These measurements characterize the practical inference cost of PREX under this configuration.

\begin{figure*}[t]
  \centering
  \includegraphics[width=0.9\linewidth]{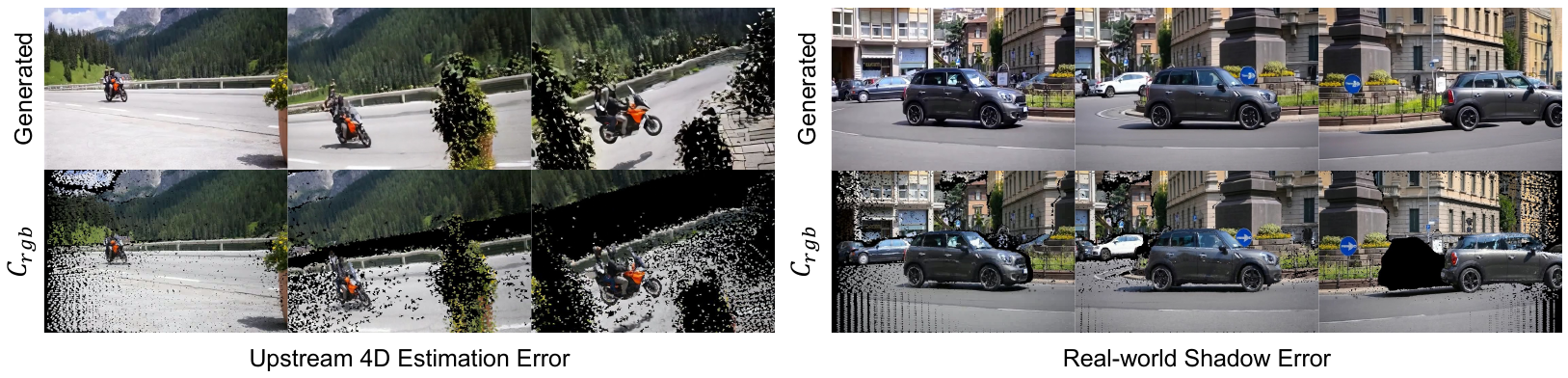}
  \caption{Qualitative demonstration on failure cases of PREX.}
  \label{fig:failure_cases}
\end{figure*}

\section{Failure cases analysis.}
Figure~\ref{fig:failure_cases} shows two representative failure cases of PREX. 
First, PREX depends on the quality of the upstream 4D estimation. Extensive missing regions or severe structural errors in the reconstructed geometry or projected appearance cue \(C^{rgb}\) can propagate to the generated video, leading to local distortion, unstable object boundaries, or inaccurate preservation. 
Second, PREX may struggle with complex real-world illumination effects, especially shadows. Since shadows are not explicitly modeled as part of the editable 4D representation, the rendered cue can contain inconsistent or incomplete shadow evidence after camera or object edits. As a result, the generated video may produce unnatural dark regions or fail to maintain physically consistent contact shadows. 
These cases suggest that improving upstream 4D reconstruction quality and incorporating illumination-aware or shadow-aware conditioning are important directions for future work.

\section{Broader Impact}
\label{app:broader_impact}

PREX targets faithful 4D video editing, with potential positive applications in controllable content creation, film and media production, virtual production, robotics simulation, embodied AI data generation, and interactive 4D scene authoring. 
By explicitly separating source-supported preservation regions from newly revealed and expanded regions, the method may also help make video-editing failures more diagnosable and measurable, rather than relying only on holistic video-quality scores.

At the same time, improved video editing methods can be misused to create misleading or deceptive visual content. 
Potential negative impacts include generating manipulated videos for misinformation, impersonation, harassment, or unauthorized alteration of real-world scenes and people. 

To mitigate these risks, we recommend that released models and demos include clear usage restrictions prohibiting impersonation, non-consensual editing of people, deceptive political or news content, and other malicious uses. 
We also recommend provenance-preserving release practices, such as retaining metadata for generated videos, supporting watermarking or disclosure mechanisms when possible, and encouraging users to clearly label AI-edited outputs. 
For benchmark release, we will avoid distributing unsafe, explicit, private, or personally sensitive content, and will only release assets in accordance with the licenses and terms of the underlying datasets.



\end{document}